\documentclass[acmsmall,screen,manuscript]{acmart}
\AtBeginDocument{%
  \providecommand\BibTeX{{%
    \normalfont B\kern-0.5em{\scshape i\kern-0.25em b}\kern-0.8em\TeX}}}
\newtheorem*{theorem*}{Comparison Principle}

\setcopyright{acmcopyright}
\copyrightyear{2018}
\acmYear{2018}
\acmDOI{XXXXXXX.XXXXXXX}

\acmJournal{TOIS}
\acmVolume{37}
\acmNumber{4}
\acmArticle{111}
\acmMonth{8}

\usepackage{algorithm}
\usepackage{algpseudocode}
\usepackage{bbm}
\usepackage{bm}
\usepackage{multirow}
\usepackage{subfigure}
\usepackage{colortbl}
\usepackage{xcolor}
\usepackage{graphicx}

\begin{document}

\title{Cross-Model Comparative Loss for Enhancing Neuronal Utility in Language Understanding}

\author{Yunchang Zhu}
\affiliation{%
    \institution{CAS Key Laboratory of AI Security,  Institute of Computing Technology, CAS; University of Chinese Academy of Sciences}
    \city{Beijing}
    \country{China}}
\email{zycdev@gmail.com}

\author{Liang Pang}
\authornote{Corresponding author}
\affiliation{%
   \institution{CAS Key Laboratory of AI Security,  Institute of Computing Technology, CAS}
    \city{Beijing}
    \country{China}}
\email{pangliang@ict.ac.cn}

\author{Kangxi Wu}
\affiliation{%
    \institution{CAS Key Laboratory of AI Security,  Institute of Computing Technology, CAS; University of Chinese Academy of Sciences}
    \city{Beijing}
    \country{China}}
\email{wukangxi22s@ict.ac.cn}

\author{Yanyan Lan}
\affiliation{%
  \institution{Institute for AI Industry Research, Tsinghua University}
  \city{Beijing}
  \country{China}
}
\email{lanyanyan@tsinghua.edu.cn}

\author{Huawei Shen}
\affiliation{%
   \institution{CAS Key Laboratory of AI Security,  Institute of Computing Technology, CAS; University of Chinese Academy of Sciences}
  \city{Beijing}
  \country{China}}
\email{shenhuawei@ict.ac.cn}

\author{Xueqi Cheng}
\affiliation{%
   \institution{CAS Key Laboratory of AI Security,  Institute of Computing Technology, CAS; University of Chinese Academy of Sciences}
  \city{Beijing}
  \country{China}}
\email{cxq@ict.ac.cn}

\thanks{This work was supported by the  \grantsponsor{GS1}{National Key R\&D Program of China}{} (\grantnum{GS1}{2022YFB3103700}, \grantnum{GS1}{2022YFB3103704}), the \grantsponsor{GS2}{National Natural Science Foundation of China (NSFC)}{} under Grants No. \grantnum{GS2}{62276248}, \grantnum{GS2}{U21B2046}, and the \grantsponsor{GS3}{Youth Innovation Promotion Association CAS}{} under Grants No. \grantnum{GS3}{2023111}.}

\renewcommand{\shortauthors}{Zhu and Pang, et al.}

\begin{abstract}
Current natural language understanding (NLU) models have been continuously scaling up, both in terms of model size and input context, introducing more hidden and input neurons.
While this generally improves performance on average, the extra neurons do not yield a consistent improvement for all instances.
This is because some hidden neurons are redundant, and the noise mixed in input neurons tends to distract the model.
Previous work mainly focuses on extrinsically reducing low-utility neurons by additional post- or pre-processing, such as network pruning and context selection, to avoid this problem. 
Beyond that, can we make the model reduce redundant parameters and suppress input noise by intrinsically enhancing the utility of each neuron?
If a model can efficiently utilize neurons, no matter which neurons are ablated (disabled), the ablated submodel should perform no better than the original full model. 
Based on such a comparison principle between models, we propose a cross-model comparative loss for a broad range of tasks. 
Comparative loss is essentially a ranking loss on top of the task-specific losses of the full and ablated models, with the expectation that the task-specific loss of the full model is minimal.
We demonstrate the universal effectiveness of comparative loss through extensive experiments on 14 datasets from 3 distinct NLU tasks based on 5 widely used pretrained language models and find it particularly superior for models with few parameters or long input.
\end{abstract}

\newcommand\blfootnote[1]{%
\begingroup
\renewcommand\thefootnote{}\footnote{#1}%
\addtocounter{footnote}{-1}%
\endgroup
}
\blfootnote{
This paper is an extension of the SIGIR 2022 conference paper~\cite{zhu2022lol}.
The earlier conference paper is limited to solving the query drift problem in pseudo-relevance feedback by comparing the retrieval loss using different size feedback sets.
However, the comparison principle and comparative loss are actually general and task-agnostic. 
Moreover, in addition to comparing the input of the model, we can also compare the parameters of the model.
Thus, in this work, we 
(1) provide a more general and complete formulation of the comparison principle and comparative loss, 
(2) directly use a unified comparative loss as the final loss being optimized, eliminating the need to set a weighting coefficient between the comparative regularization term and the task-specific losses,
(3) improve the previous comparison method that compares inputs with different context sizes, and propose an alternative dropout-based comparison method to improve the utility of the parameters to the model, and 
(4) apply the comparative loss to more tasks and models and empirically demonstrate its universal effectiveness.
}
\footrule




\maketitle

\section{Introduction}
Natural language understanding (NLU) has been pushed a remarkable step forward by deep neural models.
To further enhance the performance of deep models, enlarging model size~\cite{brutzkus2019,kaplan2020,brown2020gpt3,raffel2020t5} and input context~\cite{wang2022reina,borgeaud2022retro,izacard2022atlas} are two conventional and effective ways, where the former introduces more hidden neurons and the latter brings more input neurons.
Although neural models with more hidden or input neurons have higher accuracy on average, large-scale models do not always beat small models.
For example, on one hand, many network pruning methods have shown that compressed models with significantly reduced parameters (neuron connections) can maintain accuracy~\cite{lecun1989,han2015,liu2017} and even improve generalization~\cite{bartoldson2020}, 
\citet{meyes2019} find that ablation of neurons can consistently improve performance in some specific classes, 
and \citet{zhong2021} empirically demonstrate that larger language models indeed perform worse on a non-negligible fraction of instances.
These phenomena indicate that some hidden neurons in the currently trained model are dispensable or even obstructive.
On the other hand, much of the work on question answering~\cite{clark2018,yang2018hotpotqa} and query understanding~\cite{mitra1998,zighelnic2008,collins-thompson2009} has noted that feeding more contextual information is more likely to distract the model and hurt performance.
This is not surprising, as more input neurons not only mean more relevant features but are also likely to introduce more noise that interferes with the model.
Similar to network pruning that cuts out inefficient parameters through post-processing, many context selection methods~\cite{min2018,tu2020,zheng2020bertqe,dua2021} trim off noisy segments from the input context by pre-processing. 
In essence, both network pruning and context selection reduce inefficient hidden or input neurons through additional processing. 
However, apart from extrinsically reducing inefficient neurons, can we intrinsically improve the utility of neurons during model training?

\begin{figure}[!t]
\centering
\subfigure[]{
    \includegraphics[width=0.23\textwidth]{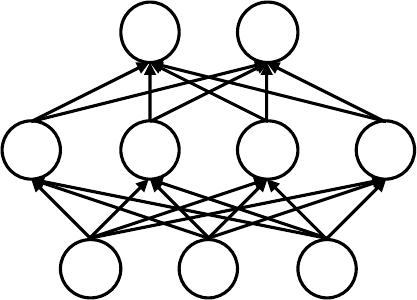}
    \label{fig:net00}
}
\subfigure[]{
    \includegraphics[width=0.23\textwidth]{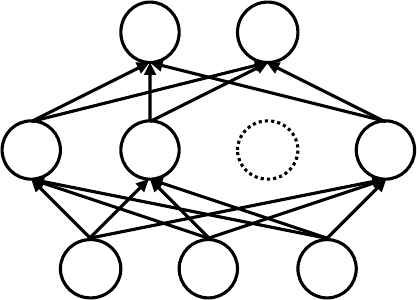}
    \label{fig:net01}
}
\subfigure[]{
    \includegraphics[width=0.23\textwidth]{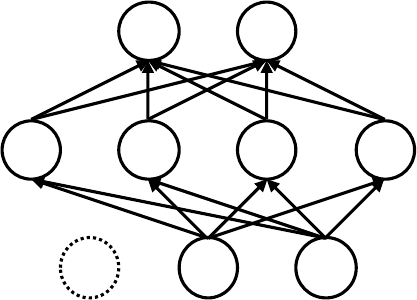}
    \label{fig:net10}
}
\subfigure[]{
    \includegraphics[width=0.23\textwidth]{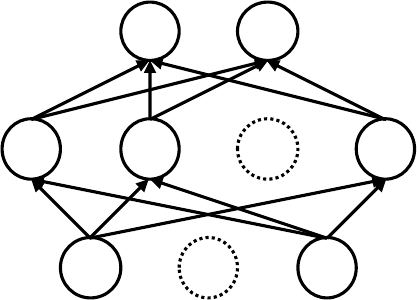}
    \label{fig:net11}
}
\caption{
An illustration of a full neural model (a) and its ablated models (b, c, and d), where a hidden neuron is ablated in (b), an input neuron is ablated in (c), and (d) additionally ablate another input neuron based on (b).
According to the comparison principle, if the full model (a) is an efficient model, the comparative relation between the task-specific losses obtained by these models should be (a) $\le$ (b), (c), (d).
If the ablated model (b) is also efficient in its parameter space, then their comparative relation can be further written as (a) $\le$ (b) $\le$ (d).
Note that (b, c) and (c, d) are two non-comparable model pairs. This is because the ablated model (c) is not a submodel of (b) and (d), and vice versa.
}
\label{fig:nets}
\end{figure}

Imagine an ideal neural network in which all its neurons should be able to cooperate efficiently to maximize the utility of each neuron. 
If a fraction of the input or hidden neurons in this network are ablated\footnote{The output value of the neuron is set to 0, which is equivalent to all the connection weights to and from this neuron being set to 0.} (disabling partial input context or model parameters), the ablated submodel is not supposed to perform better, even if the ablated neurons are noisy.
This is because an efficient\footnote{In this work, ``efficient'' refers specifically to the high utility of neurons.} model should have already suppressed these noises.
Following this intuition, we can roughly find a \textbf{comparison principle} between the original full model and its ablated model: \textbf{the fewer neurons are ablated in the model, the better the model should perform}.
During training, we can use task-specific losses as a proxy for model performance on training samples, with lower task-specific losses implying better performance.
For example, the task-specific loss of the efficient full model (a) in Fig.~\ref{fig:nets} is supposed to be minimal, and if the ablated model (b) is also efficient with respect to its restricted parameter space, the task-specific loss of the ablated model (d) is supposed to be greater than that of (b) because (d) ablates one more input neuron than (b).

Noting the gap between the ideal model and reality~\cite{min2018,zhong2021}, we aim to ensure this necessity (comparison principle) during the training to improve the model's utilization of neurons. 
Based on the natural comparison principle between models, we propose a cross-model comparative loss to train models without additional manual supervision.
In general, the comparative loss is a ranking loss on top of multiple task-specific losses. 
First, these task-specific losses are derived from the full neural model and several comparable ablated models whose neurons are ablated to varying degrees. 
Next, the ranking loss is a pairwise hinge loss that penalizes models that have fewer ablated neurons but larger task-specific losses. 
Concretely, if a model with fewer ablated neurons acquires a larger task-specific loss than another model with more ablated neurons, then the difference between the task-specific losses of the pair will be taken into account in the final comparative loss; otherwise the pair complies with the comparison principle and does not incur any training loss.
In this way, the comparative loss can drive the order of task-specific losses to be consistent with the order of the ablation degrees.
Through theoretical derivation, we also show that comparative loss can be viewed as a dynamic weighting of multiple task-specific losses, enabling adaptive assignment of weights depending on the performance of the full/ablated models.
 
The comparability among multiple ablated models is a fundamental prerequisite for comparative loss.
As a counterexample, although the ablated model (c) in Fig.~\ref{fig:nets} ablates less neurons than (d), they are not comparable and so no comparative loss can be applied.
To make the ablated models comparable with each other, we progressively ablate the models. 
The first ablated model is obtained by performing one ablation on the basis of the full model. 
If more ablated models are needed, in each subsequent ablation step we construct a new ablated model by performing a further ablation on top of the ablated model from the previous step, which makes the newly ablated model certainly a comparable submodel of the previous ones.
We provide two alternative controlled ablation methods for each ablation step, called CmpDrop and CmpCrop.
CmpDrop ablates hidden neurons by the dropout~\cite{hinton2012dropout} technique, which is theoretically applicable to all dropout-compatible models. 
While CmpCrop ablates input neurons by cropping extraneous context segments and is theoretically applicable to all tasks that contain extraneous content in the input context.

We apply comparative loss with CmpDrop or/and CmpCrop on 14 datasets from 3 NLU tasks (text classification, question answering and query understanding) with distinct prediction types (classification, extraction and ranking) on top of 5 widely used pretrained language models (PLMs)~\cite{devlin2019bert,liu2019roberta,clark2020electra,lan2020albert,beltagy2020longformer}.
The empirical results demonstrate the effectiveness of comparative loss over state-of-the-art baselines, as well as the enhanced utility of parameters and context.
Our analysis also confirms that comparative losses can indeed more appropriately weight multiple task-specific losses, as indicated by our derivation.
By exploring different comparison strategies, we observe that comparing the models ablated by first CmpCrop and then CmpDrop can bring the greatest improvement.
Interestingly, we find that comparative loss is particularly effective for models with few parameters or long inputs. 
This may imply that comparative loss can help models with lower capacity to fit the more or longer training samples better, while models with higher capacity are inherently prone to fit less data, so comparative loss is less helpful.
Moreover, we discover that different ablation methods have different effects on training, with CmpDrop helping task-specific loss to decrease to lower levels faster and CmpCrop alleviating overfitting to some extent.

The main contributions can be summarized as follows:
\begin{itemize}
    \item We propose comparative loss, a cross-model loss function based on the comparison principle between the full model and its ablated models, to improve the neuronal utility without additional human supervision.
    \item We progressively ablate the models to make multiple ablated models comparable and present two controlled ablation methods based on dropout and context cropping, applicable to a wide range of tasks and models.
    \item We theoretically show how comparative loss works and empirically demonstrate its effectiveness through experiments on 3 distinct natural language understanding tasks. We release the code and processed data at \url{https://github.com/zycdev/CmpLoss}.
\end{itemize}

\section{Preliminaries}
Before introducing our cross-model comparative loss, we review some of the concepts and notations needed afterward. 
We first introduce typical training methods for the model, followed by formalizations of network pruning and context selection methods that can further improve the model performance by removing inefficient inputs or hidden neurons. 
Finally, we elaborate on the concept of ablation, which recurs throughout the paper.

\subsection{Conventional Training}
Given a training dataset $\mathcal{D}$ for a specified task and a neural network $f$ parameterized by $\bm{\theta} \in \mathbb{R}^{|\bm{\theta}|}$, 
the training objective for each sample $(x, y) \in \mathcal{D}$ is to minimize empirical risk 
\begin{equation}
\mathcal{L}_\mathrm{emp}(x, y, \bm{\theta}) = L(y, f(x; \bm{\theta})),
\label{eq:emp}
\end{equation}
where $x$ is the input context, $y$ in output space $\mathcal{Y}$ is the label, and $L: \mathcal{Y} \times \mathcal{Y} \to \mathbb{R}_{\ge 0}$ is the task-specific loss function, $\mathbb{R}_{\ge 0}$~denoting the set of non-negative real numbers.
In NLU tasks, $x$ is typically a sequence of words, while $y$ can be a single category label for classification~\cite{pang2016, wang2019glue, xu2022}, or a pair of start and end boundaries for extraction~\cite{rajpurkar2016squad, yang2018hotpotqa, pang2019hasqa}, or a sequence of relevance levels for ranking~\cite{niu2012, nguyen2016msmarco, zhu2020l2r2, pang2020setrank}.

\subsection{Network Pruning}
After training a neural model $f(x; \bm{\theta})$, to reduce memory and computation requirements at test time, network pruning~\cite{blalock2020} entails producing a smaller model $f(x; \bm{m} \odot \bm{\theta}')$ with similar accuracy through post-hoc processing.
Here $\bm{m} \in \{0, 1\}^{|\theta|}$ is a binary mask that fixes certain pruned parameters to 0 through elementwise product $\odot$, and the parameter vector $\bm{\theta}'$ may be different from $\bm{\theta}$ because $\bm{m} \odot \bm{\theta}'$ is usually retrained from $\bm{m} \odot \bm{\theta}$ to fit the pruned network structure.

Although pruning is often viewed as a way to compress models, it has also been motivated by the desire to prevent overfitting. 
Pruning systematically removes redundant parameters and neurons that do not significantly contribute to performance and thus have much less prediction variance, which makes us reminiscent of dropout~\cite{labach2019}, another widely used technique to avoid overfitting. 
Similarly, dropout also uses a mask to disable a fraction (such as $p\%$) of parameters or neurons. 
The significant difference, though, is that the mask $\bm{m}$ in dropout is randomly sampled from a $\mathrm{Bernoulli}(1-p\%)$ distribution, rather than deterministically defined by a criterion (e.g., the bottom $p\%$ of parameters in magnitude should be masked) as in pruning. 
This in turn brings convenience: a model trained with dropout does not need to be retrained for a specific mask, because the model's neurons have already started to learn how to adapt to the absence of some neurons in the previous training.

\subsection{Context Selection}
To eliminate the noisy content in the input context $x$ and further improve the model performance, context selection selectively crops out a condensed context $x' \sqsubseteq x$ to produce the final model prediction.
In general, the model requires specialized training to fit the selected context. 
Therefore, context selection is pre-hoc processing relative to training, requiring removing the noise from the training samples in advance.
With a slight abuse of notation, here we use $x' \sqsubseteq x$ to denote that $x'$ is a condensed subsequence (possibly equal) of $x$.
In general, $x'$ is an ordered combination of segments of $x$, where the segments are usually at the sentence~\cite{min2018, pang2021}, chunk~\cite{zheng2020bertqe}, paragraph~\cite{clark2018}, or document~\cite{tu2020,dua2021} granularity.
It is worth noting that the selector for segment selection generally requires additional supervised training and needs to be run in advance of the prediction, which introduces additional computation overhead.

\subsection{Ablation}
To assess the contribution of certain components to the overall model, ablation studies investigate model behavior by removing or replacing these components in a controlled setting~\cite{cohen1988}. 
Here, in the context of machine learning, ``ablation'' refers to the removal of components of the model, which is an analogy to ablative brain surgery (removal of components of an organism) in biology~\cite{meyes2019}. 
We refer to the model after component removal as the ``ablated model'', which should continue to work. 
However, if the removed components are responsible for performance improvement, the performance of the ablated model is expected to be worse~\cite{frank2021}.

In this paper, we use ``ablation'' to refer specifically to the removal of some neurons of a neural model, i.e., to set the output of some specific neurons to zero. 
From such a neuronal perspective, network pruning and context selection can be viewed as two kinds of ablation, the former removing some low-contributing hidden neurons after training and the latter removing some low-information input neurons before training. 
However, in contrast to ablation studies that aim to investigate the role of the ablated neurons, we aim to learn to improve the utility of the ablated neurons.

\section{Methodlogy}

The primary motivation of this work is to inherently improve the utility of neurons in NLU models through a cross-model training objective, rather than post-hoc network pruning or pre-hoc context selection to eliminate inefficient neurons.
In the following, we first describe a comparison principle.
Then, we propose a novel comparative loss based on the corollary of the comparison principle and present how to train models with comparative loss by two controlled ablation methods. 
Finally, we discuss how comparative loss works.

\subsection{Comparison Principle}
For an \textit{efficient} model, we believe that all its neurons should be able to work together efficiently to maximize the utility of each neuron.
This means that each neuron should contribute to the overall model, or at least be harmless, because the cooperation of neurons is supposed to eliminate the negative effects of noise that may be produced by individual neurons. 
Thus, if we ablate some neurons, even those that produce noise, due to the missing contribution of the ablated neurons, then the ablated submodel should perform no better than the original full model, in other words, its task-specific loss should be no smaller than the original.

Formally, we define a neural model as an \textit{efficient} model if and only if it performs no weaker than any of its ablation models, and we formalize the comparison principle between an \textit{efficient} model and its ablation models as follows.

\begin{theorem*}
Suppose $f(x; \bm{\theta})$ is an \textbf{efficient} neural model for the input $x$ with respect to the parameter space $\mathbb{R}^{|\bm{\theta}|}$, let $x' \sqsubset x$ be the ablated input and $\bm{\theta}' = \bm{m} \odot \bm{\theta}$ be the ablated parameters. 
Then, for any subsequence $x'$ of $x$ whose label is still $y$, the input-ablated model $f(x'; \bm{\theta})$ should not perform better than the full model $f(x; \bm{\theta})$, 
and for any parameters $\bm{\theta}'$ masked by arbitrary $\bm{m}$, the parameter-ablated model $f(x; \bm{\theta}')$ should not perform better than $f(x; \bm{\theta})$, i.e.,
\begin{equation}
  L(y, f(x; \bm{\theta})) \le L(y, f(x'; \bm{\theta})), \forall x' \sqsubset x \text{ with } g(x') = y,
\label{eq:ie}
\end{equation}
\begin{equation}
  L(y, f(x; \bm{\theta})) \le L(y, f(x; \bm{\theta}')), \forall \bm{\theta}' = \bm{m} \odot \bm{\theta} \text{ with } \bm{m} \in \{0, 1\}^{|\bm{\theta}|},
\label{eq:pe}
\end{equation}
where $g(x')$ means the ground-truth output of $x'$.
\end{theorem*}

In the above definition, we consider that an \textit{efficient} neural model should be \textit{input-efficient} and \textit{parameter-efficient}. 
In particular, the \textit{input-efficient} property refers that the model can efficiently utilize the input neurons (words). 
If the model $f(x; \bm{\theta})$ satisfies Eq.~\eqref{eq:ie}, we say that $f(\cdot; \bm{\theta})$ is \textit{input-efficient} for the input $x$. 
The \textit{parameter-efficient} property refers that the model can utilize the hidden neurons efficiently. 
If the model $f(x; \bm{\theta})$ satisfies Eq.~\eqref{eq:pe}, we say $f(x; \bm{\theta})$ is \textit{parameter-efficient} for the input $x$ with respect to the parameter space $\mathbb{R}^{|\bm{\theta}|}$.
According to Eq.~\eqref{eq:pe}, we can definitely find at least one vector $\bm{\theta}$ that is \textit{parameter-efficient} for the input $x$, i.e., the zero vector and the optimal parameter vector that minimizes the empirical risk. 
If the parameter space is large enough, from those vectors \textit{parameter-efficient} for $x$, we can find some parameters that simultaneously satisfy Eq.~\eqref{eq:ie}, i.e, the \textit{input-efficient} property. 
That is, if the parameter space $\mathbb{R}^{|\bm{\theta}|}$ is large enough, there exists at least one parameter vector $\bm{\theta}$ that makes the model $f(x; \bm{\theta})$ \textit{efficient} for $x$.
Specially, if all activation functions in the neural model $f$ have zero output values for zero, then $f(x'; \bm{0}) = f(x; \bm{0})\ \forall x' \sqsubset x$, and hence the parameter vector $\bm{\theta} = \bm{0}$ is \textit{efficient} for any $x$.

Notably, we restrict the ablated input $x'$ for comparison to only those subsequences whose ground-truth output $g(x')$ remains unchanged, i.e., $g(x') = g(x) = y$.
This is because ablation may remove some key information from the original input $x$, such as the trigger words in the classification, resulting in an unknown change in the label $y$. 
In this unusual case, $L(y, f(x'; \bm{\theta}))$ will no longer be a reasonable proxy of the performance of the input-ablated model, so it makes no sense to compare it to the task-specific loss of the original model.
For example, in binary classification ($y \in \{0, 1\}$), for the original input $x$, $f(x; \bm{\theta})$ predicts the correct category $y$ with low confidence, whereas for the ablated input $x'$ whose category label changes to $g(x')=1-y$, $f(x'; \bm{\theta})$ predicts $1-y$ with high confidence. 
Even though $L(y, f(x; \bm{\theta})) \le L(y, f(x'; \bm{\theta}))$, the input-ablated model $f(x'; \bm{\theta})$ actually outperforms the original $f(x; \bm{\theta})$, i.e., $L(y, f(x; \bm{\theta})) \ge L(1-y, f(x'; \bm{\theta}))$, and we cannot consider $f(\cdot; \bm{\theta})$ to be input-efficient for $x$.
Although we can use $L(g(x'), f(x'; \bm{\theta}))$ as a performance proxy for the input-ablated model in Eq.~\eqref{eq:ie}, in practice it is difficult to know how the labels of the ablated inputs will change, so we try to avoid such label-changing scenarios.
For the sake of concision, from here on, we default the ablation of the input context does not change the output label if not otherwise specified, i.e., $g(x')=y$.

Further, we can ablate a full model $f(x^{(0)}; \bm{\theta}^{(0)})$ multiple ($c$) times, but are these ablated models $\{f(x^{(i)}; \bm{\theta}^{(i)})\}_{i=1}^{c}$ comparable to each other? 
The comparison principle only points out the comparative relation between an \textit{efficient} model and any of its ablated models, and cannot be directly applied to multiple ablated models.
However, if we assume that these ablated models are constructed step by step, i.e., each ablated model $f(x^{(j)}; \bm{\theta}^{(j)})$ is obtained by progressively ablating the input ($x^{(j)} \sqsubset x^{(j-1)}$) x or parameters ($\bm{\theta}^{(j)} = \bm{m}^{(j)} \odot \bm{\theta}^{(j-1)}$) based on its previous model $f(x^{(j-1)}; \bm{\theta}^{(j-1)})$, then $f(x^{(j)}; \bm{\theta}^{(j)})$ can be considered as an ablated model of all its ancestor models $\{f(x^{(i)}; \bm{\theta}^{(i)})\}_{i=0}^{j-1}$.
For simplicity, we abbreviate their task-specific losses as $l^{(i)} = L(y, f(x^{(i)}; \bm{\theta}^{(i)}))$.
Furthermore, if all $\{f(x^{(i)}; \bm{\theta}^{(i)})\}_{i=0}^{c-1}$ are simultaneously assumed to be \textit{efficient} with respect to their parameter spaces $\mathbb{R}^{\|\bm{m}^{(i)}\|_0}$~\footnote{The number of non-zeros (L0 norm) in the mask determines the number of available parameters.}, we can apply the comparison principle to compare the task-specific losses of any two models, i.e., $l^{(i)} \le l^{(j)}, \forall i < j$.

Formally, we define an \textit{efficient} model to be \textit{hereditarily efficient} if and only if its ablated models are all \textit{efficient}.
Similarly, if the parameter-ablated models of a \textit{parameter-efficient} model are all \textit{parameter-efficient}, we call this \textit{parameter-efficient} model \textit{hereditarily parameter-efficient}. 
And the \textit{hereditarily input-efficient model} is defined in the same way.
Specifically, the parameter vector $\bm{\theta} = \bm{0}$ is \textit{hereditarily parameter-efficient}, and $f(x; \bm{0})$ is also \textit{hereditarily efficient} for any $x$ if all activation functions of $f$ have zero output values for zero.

Based on the definition of the \textit{hereditarily efficient} model and the comparison principle, we can draw the following corollary.

\begin{corollary}
\label{corollary:cmp}
Suppose $f(x^{(0)}; \bm{\theta}^{(0)})$ is a \textbf{hereditarily efficient} neural model for the input $x^{(0)}$ with respect to the parameter space $\mathbb{R}^{|\bm{\theta}^{(0)}|}$, let $\{f(x^{(i)};  \bm{\theta}^{(i)})\}_{i=1}^{c}$ be its multiple progressively ablated models, where $x^{(i)} \sqsubset x^{(i-1)}$ x or $\bm{\theta}^{(i)} = \bm{m}^{(i)} \odot \bm{\theta}^{(i-1)}$.
Then, all $\{f(x^{(i)}; \bm{\theta}^{(i)})\}_{i=1}^{c-1}$ are also efficient models with respect to their parameter spaces $\mathbb{R}^{\|\bm{m}^{(i)}\|_0}$, and their task-specific losses should be monotonically non-decreasing with the degrees of ablation, i.e.,
\begin{equation}
  l^{(0)} \le l^{(1)} \cdots \le l^{(i)} \cdots \le l^{(c)}.
\label{eq:ee}
\end{equation}
\end{corollary}

In brief, the corollary describes a desirable transitive comparison between a \textit{hereditarily efficient} neural model and its ablated models, i.e., the less ablation, the better the performance.
Unfortunately, this natural property has been largely ignored before, which motivates us to exploit it to train models that utilize neurons more efficiently.

\subsection{Comparative Loss}

\begin{figure}[!t]
\centering
\includegraphics[width=0.6\linewidth]{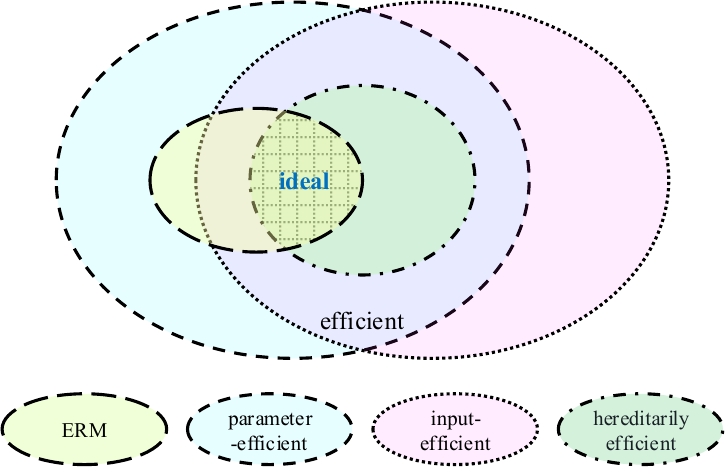}
\caption{The Venn diagram for some of the concepts in this paper. 
The \textit{empirical risk minimized (ERM)} refers to the minimization of Eq.~\eqref{eq:emp}, which is a subset of the \textit{parameter-efficient} (satisfying Eq.~\eqref{eq:pe}). 
The \textit{efficient} (intersecting purple region) model in the comparison principle, in addition to being \textit{parameter-efficient}, also needs to be \textit{input-efficient} (satisfying Eq.~\eqref{eq:ie}). 
The \textit{hereditarily efficient} model requires not only the full model to be \textit{efficient}, but also any of its ablated models to be \textit{efficient}, i.e., satisfying Eq.~\eqref{eq:ee} in Corollary~\ref{corollary:cmp}. 
The training objective of the comparative loss Eq.~\eqref{eq:cmp} is both \textit{hereditarily efficient} and \textit{ERM}, i.e., the central overlapping grid region.
}
\label{fig:veen}
\end{figure}

Based on Corollary~\ref{corollary:cmp}, we can train a \textit{hereditarily efficient} model with the objective of ordered comparative relation in Eq.~\eqref{eq:ee}.
To measure the difference from the desirable order, we can use pairwise hinge loss~\cite{herbrich2000} to evaluate the ranking of the task-specific losses of the full model and its ablated models, like $\sum_{i=0}^{c-1}\sum_{j=i+1}^{c} \max(0, l^{(i)} - l^{(j)})$. 
However, optimizing this ranking loss alone cannot guarantee that these task-specific losses are minimized, i.e., the full/ablated models may not be \textit{empirical risk minimized (ERM)}~\cite{vapnik1991erm} with respect to their parameter spaces.
To push these models to be \textit{ERM}, we introduce a special scalar $b$ as the baseline value of the task-specific loss and assume that it is derived from a dummy ablated model $f(x^{(c+1)}; \bm{\theta}^{(c+1)})$.
The dummy model is set to have the highest degree of ablation, and in principle, its task-specific loss $l^{(c+1)}$ should be the highest. 
However, to push the task-specific losses of the real models $\{f(x^{(i)},\bm{\theta}^{(i)})\}_{i=0}^{c}$ down, we usually set $l^{(c+1)}=b$ to a small value (e.g., 0) and expect all $\{l^{(i)}\}_{i=0}^{c}$ to be reduced by this target.
In this way, our comparative losses can still be written as a pairwise ranking loss, except that on top of the $c + 2$ task-specific losses,
\begin{equation}
\mathcal{L}_\mathrm{cmp}(x, y, \bm{\theta}) = \sum_{i=0}^{c}\sum_{j=i+1}^{c+1}\max\big(0, l^{(i)} - l^{(j)}\big).
\label{eq:cmp}
\end{equation}

Fig.~\ref{fig:veen} visualizes the localization (central grid region) of the ideal model of comparative loss, which is both \textit{ERM} and \textit{hereditarily efficient}. 
The \textit{hereditarily efficient} is a subset of the \textit{efficient}, and the \textit{efficient} is the intersection of the \textit{input-efficient} and \textit{parameter-efficient}.
In this light, comparative loss sets a stricter training objective than \textit{ERM}. 
When we set $c$ and $b$ to 0, Eq.~\eqref{eq:cmp} can degenerate to Eq.~\eqref{eq:emp}.
Further, the comparative loss is equivalent to
\begin{displaymath}
\sum_{l^{(i)} > b}l^{(i)} + \sum_{i=0}^{c-1}\sum_{j=i+1}^{c}\max\big(0, l^{(i)} - l^{(j)}\big),
\end{displaymath}
where the first term is to minimize the empirical risk of those not reaching the target $b$, and the second term constrains the comparative relation to pursue the full model being \textit{hereditarily efficient}.

\begin{figure*}[!t]
\centering
\includegraphics[width=0.9\textwidth]{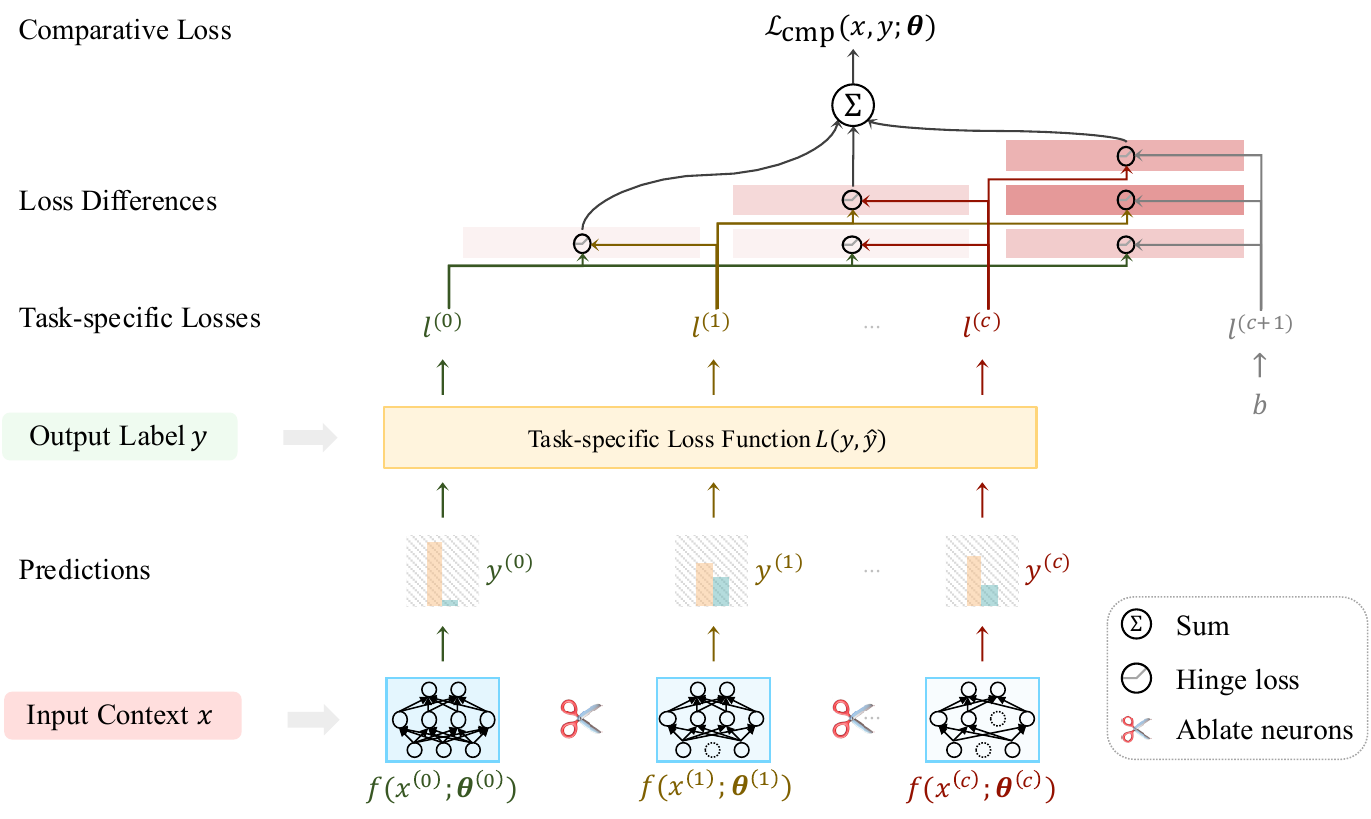}
\caption{The overview of comparative loss (best viewed in color). 
Given a data sample $(x, y)$, conventional training typically feeds the input context $x$ into the neural model to obtain the prediction $y^{(0)}$ and then just minimizes the task-specific loss $l^{(0)}$. 
In contrast, comparative loss not only progressively ablates the original model to minimize multiple task-specific losses $\{l^{(i)}\}_{i=0}^{c}$, but also constrains their comparative relation with a pairwise hinge loss.
}
\label{fig:framework}
\end{figure*}

To train using comparative loss, we first need to obtain several comparable ablated models and task-specific losses. 
As shown in Fig.~\ref{fig:framework}, we consider the original model with the input of the entire context as the full model $f(x^{(0)}; \bm{\theta}^{(0)})$. 
According to Corollary~\ref{corollary:cmp}, we progressively perform $c$-step ablation based on the full model. 
At the $i$-th ($1 \le i \le c$) ablation step, we use CmpCrop or CmpDrop to ablate a small portion of the input or hidden neurons based on the model $f(x^{(i-1)}; \bm{\theta}^{(i-1)})$ from the previous step, which makes the newly ablated model $f(x^{(i)}; \bm{\theta}^{(i)})$ comparable to all its ancestor models.
After all these models have predicted once, we have $c+1$ comparable task-specific losses. 
Together with $l^{(c+1)}=b$ from the dummy ablated model, we can calculate the final loss using Eq.~\eqref{eq:cmp}.
Using stochastic gradient descent optimization as an example, Algorithm~\ref{alg:training} illustrates the training process more formally.

\begin{algorithm}[!t]
\caption{Training with Comparative Loss}
\label{alg:training}
\begin{algorithmic}[1]
\renewcommand{\algorithmicrequire}{\textbf{Input:}}
\renewcommand{\algorithmicensure}{\textbf{Output:}}
\Require Training dataset $\mathcal{D}$, steps of ablation $c$, dropout rate $p$, baseline value of task-specific loss $b$, learning rate $\eta$.
\Ensure model parameters $\bm{\theta}$.
\State Randomly initialize model parameters $\bm{\theta}$
\While{not converged}
  \State randomly sample a data pair $(x, y) \sim \mathcal{D}$
  \State $x^{(0)} \gets x, \bm{\theta}^{(0)} \gets \bm{\theta}$ 
  \State $l^{(0)} \gets L(y, f(x^{(0)}; \bm{\theta}^{(0)}))$
  \For{$i \gets 1$ to $c$}
    \If {ablate hidden neurons}  \Comment{ablate model parameters}
      \State $\bm{\theta}^{(i)} \gets \mathrm{CmpDrop}(\bm{\theta}^{(i-1)}, p)$  
      \State $x^{(i)} \gets x^{(i-1)}$
    \Else  \Comment{ablate input context}
      \State $x^{(i)} \gets \mathrm{CmpCrop}(x^{(i-1)})$
      \State $\bm{\theta}^{(i)} \gets \bm{\theta}^{(i-1)}$
    \EndIf
    \State calculate the task-specific loss: $l^{(i)} \gets L(y, f(x^{(i)}; \bm{\theta}^{(i)}))$
  \EndFor
  \State set the dummy's task-specific loss: $l^{(c+1)} \gets b$
  \State calculate the comparative loss $\mathcal{L}_\mathrm{cmp}(x,y,\bm{\theta})$ by Eq.~\eqref{eq:cmp}
  \State update parameters: $\bm{\theta} \gets \bm{\theta} - \eta\nabla_{\bm{\theta}}\mathcal{L}_\mathrm{cmp}(x,y,\bm{\theta})$
\EndWhile
\end{algorithmic}
\end{algorithm}

CmpDrop and CmpCrop in Algorithm~\ref{alg:training} are the two alternative ablation methods we present for each ablation step, the former for ablating the parameters (hidden neurons) and the latter for ablating the input context (input neurons).
They both randomly ablate neurons in a controlled manner on top of the previous model, which allows the coverage of all potential ablated models without retraining each ablated model. 
This is because the randomly ablated models are jointly trained and adapt to the absence of some neurons during the training process.
As for which one to use at each ablation step can be specific to the model and task dataset.
Ideally, CmpDrop can be used as long as the model is dropout compatible, and CmpCrop can be used as long as the input context of the task contains dispensable segments.
Below we will introduce CmpDrop and CmpCrop in detail.

\subsubsection{CmpDrop: Ablate Parameters by Dropout}
Dropout randomly disables each neuron with probability $p$, which coincides with our need to randomly ablate hidden neurons.
To obtain a model $f(\cdot; \bm{\theta}^{(i)})$ with more ablated parameters, instead of simply applying a larger dropout rate on the original model $f(\cdot; \bm{\theta}^{(0)})$, we ablate the surviving neurons from the previous ablated model $f(\cdot; \bm{\theta}^{(i-1)})$ with probability $p$ consistently.
Specifically, the output values of those dropped neurons are set to zeros, and the output values of the surviving neurons are scaled by $1/(1-p)$ to ensure consistency with the expected output value of a neuron in all full/ablated models~\cite{labach2019}.
This is equivalent to applying a mask with scaling\footnote{Slightly different from the binary mask in the comparison principle, we incorporate the scaling factors together into the mask in order to still express the parameter ablation concisely by $\bm{\theta}^{(i)} = \bm{m}^{(i)} \odot \bm{\theta}^{(i-1)}$.} $\bm{m}^{(i)} \in \{0, 1, 1/(1-p)\}^{|\bm{\theta}|}$ to the previous parameters $\bm{\theta}^{(i-1)}$ to obtain the ablated parameters $\bm{\theta}^{(i)} = \bm{m}^{(i)} \odot \bm{\theta}^{(i-1)}$. 
Each element in $\bm{m}^{(i)}$ corresponds to the scaling factor of each parameter,
\begin{displaymath}
m_k^{(i)} = 
\begin{cases}
1,  & \text{if no dropout in $\theta_k$'s layer;} \\
\frac{1}{1-p}, & \text{if neurons at both ends of $\theta_k$ survive;} \\
0, & \text{otherwise.}
\end{cases}
\end{displaymath}
For the third case in the above equation, once a neuron is newly ablated, all connection parameters from and to it are set to zero; in addition, parameters that have been ablated by $\bm{m}^{(i-1)}$ are still set to zero.

In practice, we can leverage the existing dropout to implement it. 
However, for the comparability of the ablated models, we must use the same random seed and the same state of the random number generator in each CmpDrop. 
In this way, assuming that the current ablation step is the $n$-th execution of CmpDrop, we can simply run the model with the dropout rate of $1-(1-p)^n$.

\subsubsection{CmpCrop: Ablate Input by Cropping}\label{sec:cc}
Given an input context $x$, CmpCrop aims to crop out a condensed context $x'$ that does not change the original ground-truth output, i.e. $x' \sqsubset x$ and $g(x') = g(x) = y$. 
Assume that we know the minimum support context $x^{\star}$ for $x$ at training time, i.e., $\forall x' \sqsupseteq x^{\star}\ g(x') = g(x^{\star})$ and $\forall x' \sqsubset x^{\star}\ g(x') \ne g(x^{\star})$. 
Then, CmpCrop can produce a streamlined context by randomly cropping out several insignificant segments from the non-support context $x \setminus x^{\star}$. 
In this way, the trimmed streamlined context is sure to contain the minimum support context, so the ground-truth output does not change.

In practice, to use CmpCrop, we must ensure that enough insignificant segments are set aside in the original context $x^{(0)}$ for cropping. 
The segments can be of document, paragraph or sentence granularity.
For example, in question answering, an insignificant segment can be any retrieved paragraph that does not affect the answer to the question.
If the dataset does not annotate the minimal support context, we can manually inject a few extraneous noise segments into $x^{(0)}$.

\subsection{Discussion}\label{sec:disc}
Further deriving Eq.~\eqref{eq:cmp}, we find that comparative loss can be viewed as a dynamic weighting of multiple task-specific losses.
In particular, the loss can be rewritten as follows,
\begin{equation}
\begin{aligned}
\mathcal{L}_\mathrm{cmp} &= \sum_{i=0}^{c}\sum_{j=i+1}^{c+1}\max\big(0, l^{(i)} - l^{(j)}\big)
\\ &= \sum_{i=0}^{c}\sum_{j=i+1}^{c+1}\mathbbm{1}_{l^{(i)}>l^{(j)}} \cdot l^{(i)} - \sum_{i=0}^{c}\sum_{j=i+1}^{c+1}\mathbbm{1}_{l^{(i)}>l^{(j)}} \cdot l^{(j)}
\\ &= \sum_{i=0}^{c+1}\sum_{j=i+1}^{c+1}\mathbbm{1}_{l^{(i)}>l^{(j)}} \cdot l^{(i)} - \sum_{j=0}^{c}\sum_{i=j+1}^{c+1}\mathbbm{1}_{l^{(j)}>l^{(i)}} \cdot l^{(i)}
\\ &= \sum_{i=0}^{c+1}\sum_{j=i+1}^{c+1}\mathbbm{1}_{l^{(i)}>l^{(j)}} \cdot l^{(i)} - \sum_{i=1}^{c+1}\sum_{j=0}^{i-1}\mathbbm{1}_{l^{(j)}>l^{(i)}} \cdot l^{(i)}
\\ &= \sum_{i=0}^{c+1}\sum_{j=i+1}^{c+1}\mathbbm{1}_{l^{(i)}>l^{(j)}} \cdot l^{(i)} - \sum_{i=0}^{c+1}\sum_{j=0}^{i-1}\mathbbm{1}_{l^{(j)}>l^{(i)}} \cdot l^{(i)}
\\ &= \sum_{i=0}^{c+1}\left[ \sum_{j=i+1}^{c+1}\mathbbm{1}_{l^{(i)}>l^{(j)}} \cdot l^{(i)} - \sum_{j=0}^{i-1}\mathbbm{1}_{l^{(j)}>l^{(i)}} \cdot l^{(i)} \right]
\\ &= \sum_{i=0}^{c+1} l^{(i)} \cdot \left[ \sum_{j=i+1}^{c+1}\mathbbm{1}_{l^{(i)}>l^{(j)}} - \sum_{j=0}^{i-1}\mathbbm{1}_{l^{(i)}<l^{(j)}} \right]
\\ &= \sum_{i=0}^{c+1} \sum_{j \neq i}\mathrm{CMP}(i, j, l^{(i)}, l^{(j)}) \cdot l^{(i)},
\end{aligned}
\label{eq:cmp1}
\end{equation}
where $\mathbbm{1}_C$ is an indicator function equal to 1 if condition $C$ is true and 0 otherwise, and the $\mathrm{CMP}$ function determines whether model $f(x^{(i)}; \bm{\theta}^{(i)})$ complies with the comparison principle compared to $f(x^{(j)}; \bm{\theta}^{(j)})$ and adjusts the weight of $l^{(i)}$.
There are two cases of non-compliance: for the case where $f(x^{(i)}; \bm{\theta}^{(i)})$ is less ablated ($i < j$) but more loss is obtained, we increase the weight of $l^{(i)}$; for the case where $f(x^{(i)}; \bm{\theta}^{(i)})$ is more ablated ($i > j$) but less loss is obtained, we decrease the weight of $l^{(i)}$.
Formally, the $\mathrm{CMP}$ function can be written as
\begin{displaymath}
\mathrm{CMP}(i, j, l^{(i)}, l^{(j)}) = 
\begin{cases} 
1,  & \text{if $i < j$ and $l^{(i)} > l^{(j)}$;} \\
-1, & \text{if $i > j$ and $l^{(i)} < l^{(j)}$;} \\
0, & \text{otherwise.}
\end{cases}
\end{displaymath}
Here we can notice that for a pair of models that do not conform to the comparison principle, we increase (+1) the weight of the task-specific loss of the model that is ablated less and equally decrease (-1) the weight of the loss of the model that is ablated more.
Thus, let $\alpha^{(i)} = \sum_{j \neq i}\mathrm{CMP}(i, j, l^{(i)}, l^{(j)})$ denote the weight of $l^{(i)}$, then the sum of the weights of all task-specific losses (including the dummy one) is 0, i.e., $\sum_{i=0}^{c}\alpha^{(i)}=-\alpha^{(c+1)}=\sum_{i=0}^{c}\mathbbm{1}_{l^{(i)}>b}$.
Since $l^{(c+1)}=b$ is a constant, Eq.~\eqref{eq:cmp1} is also equivalent to $\sum_{i=0}^{c}\alpha^{(i)}l^{(i)}$, i.e., the total weight equals the number of task-specific losses worse than the virtual baseline $b$, and is adaptively assigned to the $c+1$ losses according to their performance.
In this way, poorly performing full/ablated models will be more heavily optimized.
And we empirically compare other heuristic weighting strategies in $\S$\ref{sec:weighting}.

For parameter ablation, in addition to being able to weight each task-specific loss differentially, the comparative loss with CmpDrop can also differentially calculate the gradients of the parameters in different parts. 
According to Eq.~\eqref{eq:cmp1}, comparative loss is equal to the sum of all differences of task-specific loss pairs that violate the comparison principle, i.e, $\sum_{i<j\,\land\,l^{(i)}>l^{(j)}}\big(l^{(i)}-l^{(j)}\big)$, so we can analyze the gradient of comparative loss from the difference of each task-specific loss pair. 
For ease of illustration, we take the original model $f(x; \bm{\theta})$ and a model $f(x'; \bm{\theta}')$ whose parameters have been ablated $n$ times as an example, and other model pairs with different parameters are similar.
Assume that the original parameters $\bm{\theta}=(\bm{u},\bm{v},\bm{w})$ and the ablated parameters $\bm{\theta}'=(\bm{u}',\bm{v}',\bm{w}')=(\bm{u},\bm{v}/(1-p)^{n},\bm{0})$, where $\bm{u}$ is the parameters from the layers without dropout, $\bm{w}$ is the parameters ablated by $n$ times CmpDrop, and $\bm{v}'$ is the scaled parameters surviving from $n$ times dropout.
Then, if their task-specific losses violate the comparison principle, i.e., $l>l'$, the gradient of the comparative loss contributed by this model pair is
\begin{displaymath}
\begin{aligned}
\nabla_{\bm{\theta}}(l-l') &= (\nabla_{\bm{u}}(l-l'), \nabla_{\bm{v}}(l-l'), \nabla_{\bm{w}}l).
\end{aligned}
\end{displaymath}
We can see that the comparative loss with respect to $\bm{w}$ is higher than the comparative loss with respect to the other parameters.
This is intuitive because the model instead performs better after ablating away $\bm{w}$ indicating that the current $\bm{w}$ is inefficient, so we need to focus on updating $\bm{w}$.

In addition to the dynamic weighting perspective, comparative loss can also be considered as an ``inverse ablation study'' during training. 
This is because, in contrast to ablation studies that determine the contribution of removed components during validation, comparative loss believes that the ablated neurons should contribute and optimizes parameters with this objective.

For training complexity, given a generally small number of comparisons $c$ (i.e., number of ablation steps), the overhead of computing the final comparative loss is negligibly small, and the increased computation overhead per update step comes mainly from the multiple forward and backward propagations of the models.
Specifically, the overhead of a training step using comparison loss is $1+c$ times that of conventional training for the same batch size. 
For inference complexity, however, models trained using comparative loss are the same as conventionally trained models at test time.

\section{Experiments}
To evaluate the effectiveness and generalizability of our approach for natural language understanding, we conduct experiments on 3 tasks with representative output types, including classification (8 datasets), extraction (2 datasets), and ranking (4 datasets).
Among them, the classification task requires predicting a single category for a piece of text or a text pair, the extraction task requires predicting a pair of boundary positions to extract the span between the start and end boundaries, and the ranking task requires predicting a list of relevance level to rank candidates.
Specifically, the three distinct tasks are text classification (see $\S$\ref{sec:cls}), reading comprehension (extraction, see $\S$\ref{sec:extraction}), and pseudo-relevance feedback (ranking, see $\S$\ref{sec:ranking}), respectively.
We evaluate the comparative loss with just CmpDrop in text classification and reading comprehension, the comparative loss with just CmpCrop in reading comprehension and pseudo-relevance feedback, and the comparative loss with both CmpDrop and CmpCrop in reading comprehension.
For each task, we first introduce the dataset used, then present the implementation of our models as well as the baselines, and finally show the experimental results.

Before we start each experiment, we explain some common experimental settings. 
For the baseline value $b$ of the task-specific loss in Algorithm~\ref{alg:training}, we provide two setting options. 
One is to simply set $b=0$, which is equivalent to setting an unreachable target value for all full/ablated models and thus pushing their task-specific losses to decrease. However, this results in the exposure of all training data to the full model and may aggravate overfitting.
Therefore, to reduce the times the full model is optimized, our second option is to set the baseline value to the task-specific loss of the full model, i.e., $b=l^{(0)}$.
In this way, the full model is optimized only when it performs worse than its ablated model.
In practice, we prefer setting $b=0$, and change to setting $b=l^{(0)}$ if we find that the model is prone to overfitting on the dataset.
For the dropout rate $p$ in each CmpDrop, we use the same setting as the baseline models, which is 0.1 in all our experiments.
For other conventional training hyperparameters, such as batch size and learning rate, we also keep the same as the carefully tuned baseline models if not specifically specified.
We implement our models and baseline models in PyTorch with HuggingFace Transformers~\cite{wolf2020}. 
All models are trained on Tesla V100 GPUs.
In the text classification and reading comprehension tasks, we trained each model with 5 random seeds. In the pseudo-relevance feedback task, we trained all models with a fixed random seed~42.
In the tables, the results presented as mean$_{\pm\text{standard deviation}}$ are tallied on the evaluation results of the five random seeds, otherwise the performance of the model trained with the random seed 42.
For convenience, in the tables, we use `Cmp' to represent the comparative loss and use `Drop' and `Crop' in parentheses to refer to CmpDrop and CmpCrop, respectively.

\subsection{Classification: Application to Text Classification}\label{sec:cls}
Text classification is a fundamental task in natural language understanding, which aims to assign a predefined category to a piece or a group of text.
In many text classification datasets, all segments of the input context seem to play an important role in the text category and there is almost no annotation of the minimal support context, so it is difficult for us to construct an input-ablated model by directly cropping the original input without changing the classification label.
That is, it is likely to violate the constraint that the label of the ablated input is unchanged in the comparison principle, and thus we cannot apply CmpCrop to this task.
However, many current neural classification models use dropout during training, so in this task, we only validate the comparative loss that uses just CmpDrop.

\subsubsection{Datasets}
The General Language Understanding Evaluation (GLUE) benchmark~\cite{wang2019glue} is a collection of diverse natural language understanding tasks.
Following~\cite{devlin2019bert}, we exclude the problematic WNLI set and conduct experiments on 8 datasets:
(1) Multi-Genre Natural Language Inference (\textbf{MNLI})~\cite{williams2018mnli} is a sentence pair classification task that aims to predict whether the second sentence is an entailment, contradiction, or neutral to the first one.
(2) Microsoft Research Paraphrase Corpus (\textbf{MRPC})~\cite{dolan2005mrpc} aims to predict if two sentences in the pair are semantically equivalent. 
(3) Question Natural Language Inference (\textbf{QNLI})~\cite{wang2019glue} is a binary sentence pair classification task that aims to predict whether a sentence contains the correct answer to a question.
(4) Quora Question Pairs (\textbf{QQP})~\cite{chen2018quora} is a binary sentence pair classification task that aims to predict whether two questions asked on Quora are semantically equivalent.
(5) Recognizing Textual Entailment (\textbf{RTE})~\cite{bentivogli2009rte} is a binary entailment task similar to MNLI, but with much fewer training samples.
(6) Stanford Sentiment Treebank (\textbf{SST-2})~\cite{socher2013sst2} is a binary sentence sentiment classification task consisting of sentences extracted from movie reviews.
(7) The Semantic Textual Similarity Benchmark (\textbf{STS-B})~\cite{cer2017stsb} is a sentence pair classification task that aims to determine how two sentences are semantically similar. 
(8) The Corpus of Linguistic Acceptability (\textbf{CoLA})~\cite{warstadt2019cola} is a binary sentence classification task aimed at judging whether a single English sentence conforms to linguistics. 

\subsubsection{Models \& Training}
Following R-Drop~\cite{liang2021}, another state-of-the-art training method leveraging dropout, we validate our comparative loss on the popular classification models based on PLMs\footnote{\url{https://github.com/huggingface/transformers/blob/v4.19.2/examples/pytorch/text-classification/README.md}}.
Specifically, we take BERT$_\mathrm{base}$~\cite{devlin2019bert}, RoBERTa$_\mathrm{base}$~\cite{liu2019roberta} and ALBERT$_\mathrm{base}$~\cite{lan2020albert} as our backbones to perform finetuning.
The task-specific loss is mean squared error (MSE) for STS-B and cross-entropy for other datasets.
We use different training hyperparameters for each dataset.
For baseline models and our models trained with comparative loss, we independently select the learning rate within \{1e-5, 2e-5, 3e-5, 4e-5\}, warmup rate within \{0, 0.1\}, the batch size within \{16, 24, 32\}, and the number of epochs from 2 to 5.
For our models, we tune the number of ablation steps $c$ (i.e., the number of CmpDrop) from 1 to 4.
Following the hyperparameter setup in R-Drop~\cite{liang2021}, we implement R-Drop for all backbone models as well to serve as a competitor, which performs dropout multiple times as CmpDrop does.

\subsubsection{Results}

\begin{table}[!t]
\centering
\caption{Classification performance on the development sets of GLUE language understanding benchmark.}
\label{tab:cls}
\scalebox{0.96}{
\begin{tabular}{lccccccccc} 
\toprule
Model                                          & MNLI                                      & MRPC                                      & QNLI                                      & QQP                                       & RTE                                       & SST-2                                     & STS-B                                     & CoLA                                      & Average                                      \\ 
\midrule
BERT$_\mathrm{base}$~\cite{devlin2019bert}                           & 84.2$_{\pm0.3}$           & 85.9$_{\pm0.5}$           & 91.0$_{\pm0.1}$           & 91.0$_{\pm0.1}$           & 68.2$_{\pm1.7}$           & 92.2$_{\pm0.2}$           & 88.9$_{\pm1.0}$           & 61.9$_{\pm1.1}$           & 82.92$_{\pm0.14}$            \\
~ + R-Drop~\cite{liang2021}                                     & 84.3$_{\pm0.2}$           & 86.5$_{\pm0.5}$           & 91.6$_{\pm0.1}$           & 91.4$_{\pm0.1}$           & 68.8$_{\pm1.2}$           & 92.2$_{\pm0.3}$           & 89.4$_{\pm0.8}$           & 62.9$_{\pm0.7}$           & 83.38$_{\pm0.14}$            \\
~ + Cmp ($c$ Drop)                        & \textbf{84.8}$_{\pm0.2}$  & \textbf{87.1}$_{\pm1.0}$  & \textbf{91.9}$_{\pm0.2}$  & \textbf{91.5}$_{\pm0.1}$  & \textbf{70.5}$_{\pm1.2}$  & \textbf{93.1}$_{\pm0.4}$  & \textbf{89.7}$_{\pm0.6}$  & \textbf{63.4}$_{\pm1.1}$  & \textbf{83.96}$_{\pm0.31}$   \\ 
\hline
RoBERTa$_\mathrm{base}$~\cite{liu2019roberta}     & 87.4$_{\pm0.3}$          & 89.9$_{\pm0.6}$          & 92.8$_{\pm0.2}$          & 91.4$_{\pm0.1}$          & 76.5$_{\pm0.9}$          & 95.1$_{\pm0.2}$          & 90.8$_{\pm0.1}$          & 62.7$_{\pm0.6}$          & 85.82$_{\pm0.13}$           \\
~ + R-Drop~\cite{liang2021}                  & 87.6$_{\pm0.1}$          & 90.0$_{\pm0.6}$          & 92.9$_{\pm0.1}$          & 91.5$_{\pm0.0}$          & 78.5$_{\pm1.5}$          & 95.4$_{\pm0.2}$          & 91.1$_{\pm0.1}$          & 64.0$_{\pm0.5}$          & 86.37$_{\pm0.26}$           \\
~ + Cmp ($c$ Drop)     & \textbf{88.0}$_{\pm0.1}$ & \textbf{90.5}$_{\pm0.5}$ & \textbf{93.3}$_{\pm0.1}$ & \textbf{91.9}$_{\pm0.1}$ & \textbf{79.3}$_{\pm1.2}$ & \textbf{95.5}$_{\pm0.2}$ & \textbf{91.3}$_{\pm0.1}$ & \textbf{65.4}$_{\pm0.8}$ & \textbf{86.91}$_{\pm0.20}$  \\ 
\hline
ALBERT$_\mathrm{base}$~\cite{lan2020albert}      & 85.0$_{\pm0.3}$          & 88.4$_{\pm0.6}$          & 92.1$_{\pm0.3}$          & 90.4$_{\pm0.1}$          & 77.8$_{\pm0.8}$          & 93.0$_{\pm0.4}$          & 90.9$_{\pm0.2}$          & 59.8$_{\pm1.7}$          & 84.68$_{\pm0.18}$           \\
~ + R-Drop~\cite{liang2021}                  & 85.4$_{\pm0.4}$          & 88.7$_{\pm0.8}$          & 92.1$_{\pm0.2}$          & 90.5$_{\pm0.1}$          & 78.0$_{\pm1.4}$          & 93.3$_{\pm0.3}$          & 90.9$_{\pm0.2}$          & 59.7$_{\pm0.4}$          & 84.83$_{\pm0.14}$           \\
~ + Cmp ($c$ Drop)     & \textbf{85.7}$_{\pm0.1}$ & \textbf{89.5}$_{\pm0.4}$ & \textbf{92.3}$_{\pm0.1}$ & \textbf{91.0}$_{\pm0.1}$ & \textbf{78.3}$_{\pm1.4}$ & \textbf{93.5}$_{\pm0.2}$ & \textbf{91.0}$_{\pm0.2}$ & \textbf{60.3}$_{\pm0.9}$ & \textbf{85.19}$_{\pm0.24}$  \\
\bottomrule
\end{tabular}
}
\end{table}

We present classification performance in Table~\ref{tab:cls}, where the evaluation metrics are Pearson correlation for STS-B, Matthew’s correlation for CoLA, and Accuracy for the others.
For models based on BERT$_\mathrm{base}$, we can see that our model (+ Cmp) comprehensively outperforms the well-tuned baseline BERT$_\mathrm{base}$ and achieves an improvement of 1.04 points (on average), which proves the effectiveness of comparative loss in classification tasks.
Moreover, our model trained with comparative loss also outperforms the model trained with state-of-the-art R-Drop by 0.58 points on average, which demonstrates the superiority of comparative loss.
For models based on other more advanced RoBERTa$_\mathrm{base}$ and ALBERT$_\mathrm{base}$, we can find consistent improvement. 
In addition, since ALBERT reuses parameters across multiple layers, it has the smallest boostable space for parameter utilization, which is consistent with our observation that comparative loss brings the smallest boost to ALBERT.

\subsection{Extraction: Application to Reading Comprehension}\label{sec:extraction}

Extractive reading comprehension (RC)~\cite{rajpurkar2016squad,liu2019} is an essential technical branch of question answering (QA)~\cite{hirschman2001,lin2002,chen2018,rodriguez2021,zhu2021aiso}. Given a question and a context, extractive RC aims to extract a span from the context as the predicted answer.
Current dominant RC models basically use pretrained Transformer~\cite{vaswani2017transformer} architectures, which employ dropout in many layers during finetuning. This allows us to use CmpDrop to improve the utility of the model parameters.
Additionally, the given context is usually lengthy and contains many distracting noise segments, which also allows us to use CmpCrop to improve the model's utilization of the context by randomly deleting the labeled distracting paragraphs.
Therefore, we intend to verify the effectiveness of comparative loss using CmpDrop or/and CmpCrop in this task.

\subsubsection{Datasets}
We evaluate the comparative loss using only CmpDrop on SuQAD~\cite{rajpurkar2016squad}, which contains 100K single-hop questions with 9832 for validation, and HotpotQA~\cite{yang2018hotpotqa}, which contains 113K multi-hop questions with 7405 for validation.
For HotpotQA, we consider the distractor setting, where the context of each question contains 10 paragraphs, but only 2 of them are useful for answering the question, and the rest 8 are retrieved distracting paragraphs that are relevant but do not support the answer.
This allows us to evaluate the comparative loss with CmpCrop on HotpotQA distractor.

\subsubsection{Models \& Training}
We follow simple but effective RC models based on PLMs~\cite{devlin2019bert,liu2019roberta,clark2020electra,lan2020albert,beltagy2020longformer}, which take as input a concatenation of the question and the context and use a linear layer to predict the start and end positions of the answer.
And we use cross-entropy of answer boundaries as the task-specific loss function following~\cite{devlin2019bert} and use a learning rate warmup over the first 10\% steps.
For SQuAD, we use the popular BERT~\cite{devlin2019bert}, RoBERTa~\cite{liu2019roberta}, ELECTRA~\cite{clark2020electra} and ALBERT~\cite{lan2020albert} with a maximum sequence length of 512 as the backbone, all of which have successively achieved top rankings in multiple QA benchmarks~\cite{rajpurkar2016squad,rajpurkar2018squad2,yang2018hotpotqa}.
We first tune the learning rate in range \{1e-5, 3e-5, 5e-5, 8e-5, 1e-4, 2e-4\}, batch size in \{8, 12, 32\} and number of epochs in \{1, 2, 3\} for baseline models.
Then, setting $c=2$, we take these hyperparameters along and train our models using the comparative loss with two CmpDrop.
For HotpotQA, we use the state-of-the-art Longformer~\cite{beltagy2020longformer} with a maximum sequence length of 2048 as the backbone, which is fed with the format \texttt{<s> [YES] [NO] [Q] question </s> [T] title$_1$ [P] paragraph$_1$ $\cdots$ [T] title$_{10}$ [P] paragraph$_{10}$ </s>}.
The special tokens \texttt{[YES]}/\texttt{[NO]}, \texttt{[Q]}, \texttt{[T]}, and \texttt{[P]} represent yes/no answers and the beginning of questions, titles, and paragraphs, respectively.
Similarly, we select the learning rate in \{1e-5, 3e-5\}, batch size in \{6, 9, 12\} and number of epochs in \{3, 5, 8\} for the baseline model.
We then train our models with three comparative losses respectively, the first two applying one CmpDrop/CmpCrop ($c=1$), while the third applying one CmpCrop followed by one CmpDrop ($c=2$).
Besides, inheriting common hyperparameters and searching for coefficient weights $\alpha$ in \{0.1, 0.5, 1, 1.5\}, we also implement R-Drop~\cite{liang2021} as a competitor to CmpDrop.

\subsubsection{Results}

\begin{table}[!t]
\centering
\caption{Question answering performance on the development sets of SQuAD and HotpotQA distractor. 
The results with ${\dagger}$ are inquired from the authors of its paper.}
\label{tab:qa}
\begin{tabular}{lcc} 
\toprule
Model                                                            & EM                                       & F1                                        \\ 
\midrule
\multicolumn{3}{c}{SQuAD}                                                                                                                               \\
BERT$_\mathrm{base}$~\cite{devlin2019bert}                                            & 80.8                                     & 88.5                                      \\
BERT$_\mathrm{base}$ (our implementation)                        & 81.3$_{\pm0.2}$           & 88.5$_{\pm0.1}$            \\
~ + R-Drop~\cite{liang2021}                   & 82.2$_{\pm0.1}$          & 89.1$_{\pm0.1}$           \\
~ + Cmp (2 Drop)                                                 & \textbf{82.3}$_{\pm0.2}$  & \textbf{89.3}$_{\pm0.1}$   \\ 
\hline
RoBERTa$_\mathrm{base}$~\cite{liu2019roberta} (our implementation) & 85.8$_{\pm0.1}$          & 92.2$_{\pm0.1}$           \\
~ + R-Drop~\cite{liu2019roberta}                                   & 86.4$_{\pm0.1}$          & 92.3$_{\pm0.1}$           \\
~ + Cmp (2 Drop)                              & \textbf{86.5}$_{\pm0.2}$ & \textbf{92.6}$_{\pm0.1}$  \\ 
\hline
ELECTRA$_\mathrm{base}$~\cite{clark2020electra}                                         & 84.5                                     & 90.8                                      \\
ELECTRA$_\mathrm{base}$ (our implementation)                     & 85.9$_{\pm0.3}$           & 92.3$_{\pm0.2}$            \\
~ + R-Drop~\cite{liang2021}                                   & 86.5$_{\pm0.1}$          & 92.3$_{\pm0.1}$           \\
~ + Cmp (2 Drop)                                                 & \textbf{86.6}$_{\pm0.1}$  & \textbf{92.7}$_{\pm0.1}$  \\ 
\hline
ALBERT$_\mathrm{base}$~\cite{lan2020albert}                       & 82.3                  & 89.3                   \\
ALBERT$_\mathrm{base}$ (our implementation)   & 83.6$_{\pm0.2}$          & 90.6$_{\pm0.1}$           \\
~ + R-Drop~\cite{liang2021}                                   & 83.7$_{\pm0.2}$          & 90.7$_{\pm0.2}$           \\
~ + Cmp (2 Drop)                              & \textbf{84.4}$_{\pm0.1}$ & \textbf{91.0}$_{\pm0.1}$  \\ 
\hline\hline
\multicolumn{3}{c}{HotpotQA}                                                                                                                            \\
Longformer$_\mathrm{base}^\dagger$~\cite{beltagy2020longformer}                              & 60.3                                     & 74.3                                      \\
Longformer$_\mathrm{base}$(our implementation)                   & 61.9$_{\pm0.4}$           & 75.6$_{\pm0.3}$            \\
~ + R-Drop~\cite{liang2021}                                   & 62.0$_{\pm0.2}$          & 76.0$_{\pm0.2}$           \\
~ + Cmp (1 Drop)                                                 & 63.0$_{\pm0.4}$           & 77.0$_{\pm0.4}$            \\
~ + Cmp (1 Crop)                                                 & 62.6$_{\pm0.2}$           & 76.4$_{\pm0.2}$            \\
~ + Cmp (1 Crop  1 Drop)                                         & \textbf{63.5}$_{\pm0.3}$  & \textbf{77.2}$_{\pm0.3}$  \\
\bottomrule
\end{tabular}
\end{table}

Since we focus on extraction here, we only measure the extracted answers using EM (exact match) and F1, which is a little different from the official HotpotQA setting that simultaneously evaluates the identification of support facts.
From Table~\ref{tab:qa} we can see that our implemented baseline models trained directly using the task-specific loss Eq.~\eqref{eq:emp} largely achieve better results than those reported in their original papers. 
Once trained using comparative loss Eq.~\eqref{eq:cmp} instead, our models can still significantly outperform these well-tuned baseline models even without re-searching the training hyperparameters, demonstrating the effectiveness of comparative loss on the extraction task.
Also, the consistent improvement based on the three different PLMs demonstrates the model-agnostic nature of comparative loss.
Furthermore, from the results on HotpotQA we can find that although both CmpDrop and CmpCrop deliver significant improvement, CmpCrop + CmpDrop achieves the best results, suggesting that CmpDrop and CmpCrop may bring different benefits to the trained models.

\subsection{Ranking: Application to Pseudo-Relevance Feedback}\label{sec:ranking}

Pseudo-relevance feedback (PRF)~\cite{attar1977} is an effective query understanding~\cite{chang2020qu} technique to improve ranking accuracy, which aims to alleviate the mismatch of linguistic expressions between a query and its potential relevant documents.
Given an original query $q$ and a document collection $C$, a base ranking model returns a ranked list $D = (d_1, d_2, \cdots, d_{|D|})$.
Let $D_{\le k}$ denote the feedback set containing the top $k$ documents, where $k$ is usual referred to as the PRF depth.
The goal of PRF is to reformulate the original query $q$ into a new representation $q^{(k)}$ using the query-relevant information in $D_{\le k}$, i.e., $q^{(k)}=f((q, D_{\le k}); \bm{\theta})$, where $q^{(k)}$ is expected to yield better ranking results.
Although PRF methods do usually improve ranking performance on average~\cite{clinchant2013}, individual reformulated queries inevitably suffer from query drift~\cite{mitra1998,zighelnic2008} due to the objectively present noise in the feedback set, causing them to be inferior to the original ones.
Therefore, we can use comparative loss with CmpCrop to train PRF models to suppress the extra increased noise by comparing the effect of queries reformulated using feedback sets with different PRF depths.

\subsubsection{Datasets}
We conduct experiments on MS MARCO passage~\cite{nguyen2016msmarco} collection, which consists of 8.8M English passages collected from the search results of Bing's 1M real-world queries.
The Train set of MS MARCO contains 530K queries (about 1.1 relevant passages per query on average), the Dev set contains 6980 queries, and the online Eval set contains 6837 queries.
Apart from these, we also consider TREC DL 2019~\cite{craswell2020}, TREC DL 2020~\cite{craswell2021}, and DL-HARD~\cite{mackie2021}, three offline evaluation benchmarks based on the MS MARCO passage collection, which contain 43, 54, and 50 fine-grained (relevance grades from 0 to 3) labeled queries, respectively. 
Among them, DL-HARD~\cite{mackie2021} is a recent evaluation benchmark focusing on complex queries.
We use MS MARCO Train set to train models, and evaluate trained models on the MS MARCO Dev set to tune hyperparameters and select model checkpoints.
The selected models are finally evaluated on the online MS MARCO Eval\footnote{\url{https://microsoft.github.io/MSMARCO-Passage-Ranking-Submissions/leaderboard/}} and three other offline benchmarks.

\subsubsection{Models \& Training}
We carry out PRF experiments on two base retrieval models, ANCE~\cite{xiong2020ance} (dense retrieval) and uniCOIL~\cite{lin2021unicoil} (sparse retrieval), respectively. 
For their PRF models, we do not explicitly modify the query text, but directly generate a new query vector for retrieval following the current state-of-the-art method ANCE-PRF~\cite{yu2021anceprf}.
This allows us to directly optimize the retrieval of reformulated queries end-to-end with the negative log likelihood of the positive document~\cite{karpukhin2020dpr} as the task-specific loss:
\begin{displaymath}
L(\bm{q}^{(k)}) = -\log\frac{e^{\mathrm{sim}(\bm{q}^{(k)}, \bm{d}^+)}}{e^{\mathrm{sim}(\bm{q}^{(k)}, \bm{d}^+)} + \sum_{d^- \in D^-} e^{\mathrm{sim}(\bm{q}^{(k)}, \bm{d}^-)}},
\end{displaymath}
where $\bm{d}^+$ is the vector of a sampled document relevant to $q$ and $\bm{q}^{(k)}$, $\mathrm{sim}(\cdot, \cdot)$ is the dot product of two vectors, and $D^-$ is the collection of negative documents for them.
Since only vectors of queries are updated\footnote{The fixed vectors of documents are restored from the index pre-built by \url{https://github.com/castorini/pyserini}.}, we mine a lite collection (5.3M for dense retrieval and 3.7M for sparse retrieval) containing positive and hard negative documents of all training queries. 
In this way, for each query, all documents in the lite collection except its positive documents can be used as its $D^-$.
In general, our PRF model consists of an encoder, a vector projector, and a pooler.
First the original query $q$ and feedback documents in $D_{\le k}$ are concatenated in order with \texttt{[SEP]} as separator and input to the encoder to get the contextual embedding of each token.
Then, the projector maps the contextual embeddings to vectors with the same dimension as the document vectors.
Finally, all token vectors are pooled into a single query vector.
For dense retrieval, the encoder is initialized from ANCE$_\mathrm{FirstP}$\footnote{\url{https://github.com/microsoft/ANCE}}, the projector is a linear layer, and the pooler applies a layer normalization on the first vector (\texttt{[CLS]}) in the sequence, as in the previous work~\cite{yu2021anceprf}.
For sparse retrieval, the encoder and projector are initialized from BERT$_\mathrm{base}$ with the masked language model head, where the projector is an MLP with GeLU~\cite{hendrycks2016gaussian} activation and layer normalization, and the pooler is composed of a max pooling operation and an L2 normalization\footnote{We find that L2 normalization helps the model train stably, and it does not change the relevance ranking of documents to a query.}.
We finetune PRF baseline models for up to 12 epochs with a batch size of 96, a learning rate selected from \{2e-5, 1e-5, 5e-6\}, and PRF depth $k$ randomly sampled from 0 to 5 for each query.
We then finetune our PRF models using the comparative loss of $c=1$ CmpCrop for up to 6 epochs with a batch size of 48.
In this way, the maximum number of training steps for our models remains the same as the baseline models, i.e., up to 12 optimizations per original query.
Due to the large training costs of using multiple random seeds, we used paired t-test to calculate significant differences in retrieval performance.

\subsubsection{Results}

\begin{table}[!t]
\centering
\caption{Retrieval performance on benchmarks built on MS MARCO passage collection.
ANCE and uniCOIL are base retrieval models, + PRF denotes the PRF baseline model, + Cmp denotes our PRF model trained with the comparative loss of 1 CmpCrop, and superscript~$^{(k)}$ represents the PRF depth used during testing. Superscript $^*$ indicates statistically significant improvements over its PRF baseline model with $p \le 0.1$.}

\label{tab:ranking}
\scalebox{0.80}{
\begin{tabular}{lcccccccccc} 
\toprule
\multirow{2}{*}{Model}                  & \multicolumn{3}{c}{MARCO Dev}                   & MARCO Eval     & \multicolumn{2}{c}{TREC DL 2019} & \multicolumn{2}{c}{TREC DL 2020} & \multicolumn{2}{c}{DL-HARD}      \\ 
\cmidrule(lr){2-4}\cmidrule(lr){5-5}\cmidrule(lr){6-7}\cmidrule(lr){8-9}\cmidrule(lr){10-11}
                                        & NDCG@10        & MRR@10         & R@1K           & MRR@10         & NDCG@10        & R@1K             & NDCG@10        & R@1K             & NDCG@10        & R@1K            \\ 
\midrule
ANCE~\cite{xiong2020ance}               & 38.76          & 33.01          & 95.84          & 31.70          & 64.76          & 75.70            & 64.58          & 77.64            & 33.39          & 76.65           \\
~ + PRF$^{(3)}$~\cite{yu2021anceprf} & 40.10          & 34.40          & 95.90          & 33.00          & 68.10          & 79.10            & 69.50          & 81.50            & 36.50          & 76.10           \\
\quad+ Cmp$^{(3)}$ (1 Crop)      & 40.68$^*$          & 34.84$^*$          & 96.94$^*$          & -              & 68.42          & 80.10$^*$            & 69.58          & 81.77            & 35.61          & 79.39$^*$           \\
\quad+ Cmp$^{(5)}$ (1 Crop)            & \textbf{41.01}$^*$ & \textbf{35.14}$^*$ & \textbf{97.03}$^*$ & \textbf{34.17} & \textbf{69.58}$^*$ & \textbf{80.81}$^*$   & \textbf{70.44}$^*$ & \textbf{82.77}$^*$   & \textbf{37.44}$^*$ & \textbf{79.55}$^*$  \\ 
\hline
uniCOIL~\cite{lin2021unicoil}           & 41.21          & 35.13          & 95.81          & 34.42          & 70.09          & 82.83            & 67.35          & 84.42            & 35.96          & 76.85           \\
~ + PRF$^{(3)}$                      & 41.76          & 35.48          & 96.85          & -              & 69.42          & 83.32            & 69.25          & 84.44            & 36.53          & 77.48           \\
\quad+ Cmp$^{(3)}$ (1 Crop)            & \textbf{42.02}$^*$ & \textbf{35.75}$^*$ & \textbf{96.91} & \textbf{35.14} & \textbf{70.10}$^*$ & \textbf{83.58}   & \textbf{69.70}$^*$ & \textbf{84.51}   & \textbf{36.90}$^*$ & \textbf{77.67}  \\
\bottomrule
\end{tabular}
}
\end{table}

We report the official metrics (MRR@10 for MARCO and NDCG@10 for others) and Recall@1K of the models on multiple benchmarks in Table \ref{tab:ranking}.
In addition to reporting results for the best-performing PRF depths (numbers in superscript brackets), for a fair comparison with ANCE-PRF$^{(3)}$ (second row), we also present the results of ANCE-PRF + Cmp$^{(3)}$, both of which use the first 3 documents as feedback.
We can see that PRF baseline models (+ PRF) indeed generally outperform their base retrieval models, except that uniCOIL-PRF degrades by 0.67 percentage points in NDCG@10 of TREC DL 2019, which reflects the presence of query drift.
Our PRF models (+ Cmp) trained with comparative loss, however, outperform their base retrieval model across the board.
Under the same use of 3 feedback documents, our ANCE-PRF + Cmp also outperforms the published state-of-the-art ANCE-PRF~\cite{yu2021anceprf} on all metrics except NDCG@10 on DL-HARD.
Moreover, when 5 feedback documents are used, ANCE-PRF + Cmp achieves a go-ahead over ANCE-PRF on NDCG@10 of DL-HARD.
For sparse retrieval, our PRF model (+ Cmp) trained with comparative loss also surpasses the strong baseline uniCOIL-PRF implemented following ANCE-PRF.
All of these results above demonstrate the effectiveness of comparative loss on the ranking task.

\section{Analysis}
In this section, we further conduct several experiments for a more thorough analysis.
First, from the dynamic weighting perspective found in $\S$\ref{sec:disc}, we examine whether the adaptive weighting of comparative loss is more effective than other weighting strategies ($\S$\ref{sec:weighting}).
Next, we try several other comparison strategies to find some guiding experience in choosing the number of ablations and ablation methods in practice ($\S$\ref{sec:cmp-stg}).
Then, to confirm the enhancement of comparative loss on the utility of hidden and input neurons, we investigate the performance of models with different numbers of parameters ($\S$\ref{sec:model-size}) and context lengths ($\S$\ref{sec:input-size}).
Furthermore, we visualize the loss curves to find the impact of the comparative losses with different ablation methods on the task-specific loss ($\S$\ref{sec:curves}).
Finally, we show the actual training overhead of comparative loss in detail ($\S$\ref{sec:cost}).

\subsection{Effect of Weighting Strategy}\label{sec:weighting}

\begin{table}[!t]
\centering
\caption{
QA performance on the development set of HotpotQA distractor with different weighting strategies.
Cmp refers to Longformer + CmpCrop + CmpDrop that adaptively weights multiple task-specific losses through comparative loss.
The others are heuristics, where AVERAGE assigns the same weights to all task-specific losses, FIRST and LAST assign weight only to the first or last, and MAX dynamically assigns weight only to the largest one.
}
\label{tab:weighting}
\begin{tabular}{lcc} 
\toprule
Weighting Method       & EM                                      & F1                                       \\ 
\midrule
Cmp                    & \textbf{63.5}$_{\pm0.3}$ & \textbf{77.2}$_{\pm0.3}$  \\ 
\hline
AVERAGE                & 63.2$_{\pm0.3}$          & 76.7$_{\pm0.3}$           \\
FIRST                  & 62.1$_{\pm0.1}$          & 75.8$_{\pm0.1}$           \\
LAST & 61.9$_{\pm0.4}$          & 75.6$_{\pm0.3}$           \\
MAX                    & 63.1$_{\pm0.3}$          & 76.7$_{\pm0.3}$           \\
\bottomrule
\end{tabular}
\end{table}

To verify the role of comparative loss from the dynamic weighting perspective, we keep all the training settings of Longformer + CmpCrop + CmpDrop from the last row of Table~\ref{tab:qa} unchanged and replace only the weighting strategy of task-specific losses with some heuristics. 
Table~\ref{tab:weighting} shows their performance on the HotpotQA development set.
AVERAGE, FIRST and LAST are three static weighting strategies.
AVERAGE assigns equal weights to all task-specific losses, while FIRST and LAST assign weight to only the first and last task-specific loss, respectively, i.e., FIRST optimizes $l^{(0)}$ of the full model without dropout and LAST optimizes $l^{(2)}$ of the model with regular dropout rate $p$ (equivalent to the baseline Longformer in Table~\ref{tab:qa}).
MAX is another dynamic weighting strategy that assigns weight to only the largest task-specific loss.
We can see that dynamic weighting in comparative losses is significantly better than these heuristic weighting strategies, which proves that comparative loss can assign weights more appropriately.
In addition, AVERAGE is better than the latter three strategies that consider only one task-specific loss, indicating that it is beneficial to consider multiple task-specific losses. 
Moreover, although the latter three are all assigned to only one task-specific loss, MAX is better than the other two, which indicates that dynamic assignment is better than static assignment.

Notably, the FIRST that directly optimizes the full model outperforms the LAST that is trained with dropout, suggesting that the inconsistency of dropout between the training and inference stages~\cite{zolna2017} may indeed lead to underfitting of the full model.
And the fact that Cmp far outperforms FIRST and LAST indicates that comparative loss can automatically strike a balance between ensuring training-inference consistency and preventing overfitting.

\subsection{Effect of Comparison Strategy}\label{sec:cmp-stg}

\begin{table}[!t]
\centering
\caption{QA performance on the development set of HotpotQA distractor with different comparison strategies. $c$ is the number of ablation steps. x 2 indicates that an ablation method is repeated twice, and $A+B$ means that $A$ is used followed by $B$.}
\label{tab:strategy}
\begin{tabular}{llcc} 
\toprule
$c$                & Ablation Order    & EM            & F1             \\ 
\midrule
\multirow{2}{*}{1} & CmpDrop           & \textbf{63.1} & \textbf{77.0}  \\
                   & CmpCrop           & 63.1          & 76.8           \\ 
\hline
\multirow{4}{*}{2} & CmpDrop x 2       & 63.4          & 77.1           \\
                   & CmpCrop x 2       & 63.0          & 76.7           \\
                   & CmpDrop + CmpCrop & 63.2          & 76.8           \\
                   & CmpCrop + CmpDrop & \textbf{63.6} & \textbf{77.4}  \\
\bottomrule
\end{tabular}
\end{table}

To study the impact of comparison strategies, i.e., how many ablation steps we should use for comparison and which ablation method we should choose at each step, we try a variety of comparison strategies on HopotQA with different numbers of comparisons and ablation orders.
As shown in Table~\ref{tab:strategy}, the results are not significantly further improved when we repeat CmpDrop/CmpCrop twice, but the results are further improved when we apply CmpCrop first and then CmpDrop. 
This indicates that comparing multiple models ablated by the same method, i.e., encouraging the model be either \textit{hereditarily input-efficient} or \textit{hereditarily parameter-efficient}, seems to have little effect on the performance of the full model, but the successive use of two different ablation methods, i.e., encouraging the model be \textit{efficient} (both \textit{input-efficient} and \textit{parameter-efficient}), is helpful.
However, applying CmpDrop followed by CmpCrop did not perform as well as applying CmpDrop only, suggesting that the order of the ablation methods is important and perhaps the ablation should be done in the order of the information flow in the model.

\begin{figure}[!t]
\centering
\includegraphics[width=0.6\textwidth]{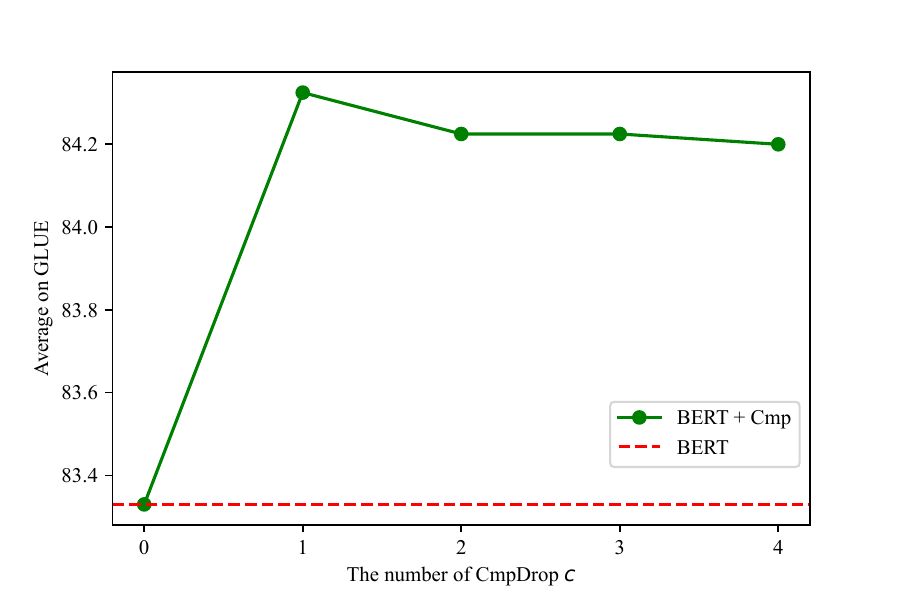}
\caption{Average results on eight GLUE datasets as the number of ablation steps changes.}
\label{fig:num-drop}
\end{figure}

To further confirm the influence of the number of ablation steps $c$, we show in Fig.~\ref{fig:num-drop} the relationship between the model's Average metric over the eight GLUE datasets and the number of ablations.
We can find little difference in the average performance of the models trained with different numbers of CmpDrop, with the model trained with one CmpDrop performing significantly best mainly because its huge advantage on two of the datasets pulls up the average.
Therefore, if there is no extreme demand for performance, we usually do not need to tune the hyperparameter $c$.

\subsection{Effect of Model Parameters}\label{sec:model-size}

\begin{table}
\centering
\caption{Evaluation results of baselines with different model sizes and initializations on the SQuAD development set (EM/F1), and relative gains of our models trained using comparative loss with CmpDrop over baselines.}
\label{tab:backbone}
\scalebox{0.93}{
\begin{tabular}{lcccccc} 
\toprule
\multirow{2}{*}{\# Parameter} & \multicolumn{2}{c}{BERT}                                                                & \multicolumn{2}{c}{ELECTRA}                                                             & \multicolumn{2}{c}{RoBERTa}                                             \\ 
\cmidrule(l){2-7}
                                & Baseline                                    & Gain (\%)                                 & Baseline                                    & Gain (\%)                                 & Baseline                  & Gain (\%)               \\ 
\midrule
Tiny: 4M                      & 41.5$_{\pm0.5}$/54.2$_{\pm0.3}$ & 2.2$_{\pm1.4}$/1.7$_{\pm1.0}$ & -                                           & -                                         & -                         & -                        \\
Small: 14M                    & 74.2$_{\pm0.6}$/82.7$_{\pm0.5}$ & 2.2$_{\pm0.8}$/1.6$_{\pm0.6}$ & 78.1$_{\pm0.2}$/85.9$_{\pm0.1}$ & 1.6$_{\pm0.5}$/1.2$_{\pm0.3}$ & -                         & -                        \\
Medium: 42M                   & 78.0$_{\pm0.4}$/85.8$_{\pm0.2}$ & 1.6$_{\pm0.7}$/1.3$_{\pm0.4}$ & -                                           & -                                         & -                         & -                        \\
Base: 110M                    & 81.3$_{\pm0.2}$/88.5$_{\pm0.1}$ & 1.2$_{\pm0.2}$/0.9$_{\pm0.2}$ & 85.9$_{\pm0.3}$/92.3$_{\pm0.2}$ & 0.8$_{\pm0.4}$/0.5$_{\pm0.3}$ & 85.8$_{\pm0.1}$/92.2$_{\pm0.1}$ & 0.8$_{\pm0.2}$/0.5$_{\pm0.1}$  \\
Large: 335M                   & 83.9$_{\pm0.2}$/90.8$_{\pm0.1}$ & 1.2$_{\pm0.2}$/0.7$_{\pm0.1}$ & 89.0$_{\pm0.1}$/94.7$_{\pm0.0}$ & 0.7$_{\pm0.2}$/0.3$_{\pm0.0}$ & 89.0$_{\pm0.3}$/94.7$_{\pm0.0}$ & 0.7$_{\pm0.3}$/0.3$_{\pm0.1}$  \\
\bottomrule
\end{tabular}
}
\end{table}

To investigate the impact of model parameters, we explore the application of the comparative loss with CmpDrop on different-sized versions of BERT, RoBERTa and ELECTRA.
From Table~\ref{tab:backbone} we can see that the comparative loss with CmpDrop achieves a consistent improvement over the baselines based on these backbone models, which indicates that the comparative loss can improve model performance by increasing parameter utility without increasing the number of parameters.
Moreover, except for the one outlier of BERT$_\mathrm{Medium}$, we can roughly find that the less the model parameters, the greater the relative gain from comparative loss. 
This is reasonable because the individual hidden neurons in a model with lower capacity play a larger role, so the improvement in the utility of hidden neurons can be more reflected in the final performance.
Whereas for a model of higher capacity, it is easier to fit less training data, i.e., its task-specific loss is already low, so comparative loss has less room to play in reducing task-specific loss further.
In addition, we observe that the boost to BERT from the comparative loss with CmpDrop is generally higher compared to RoBERTa and ELECTRA with more complicated pretraining, suggesting that the comparative loss helps the model escape from local optima due to parameter initialization.

\subsection{Effect of Input Context}\label{sec:input-size}

\begin{figure}[!t]
\centering
\subfigure[]{
    \includegraphics[width=0.44\textwidth]{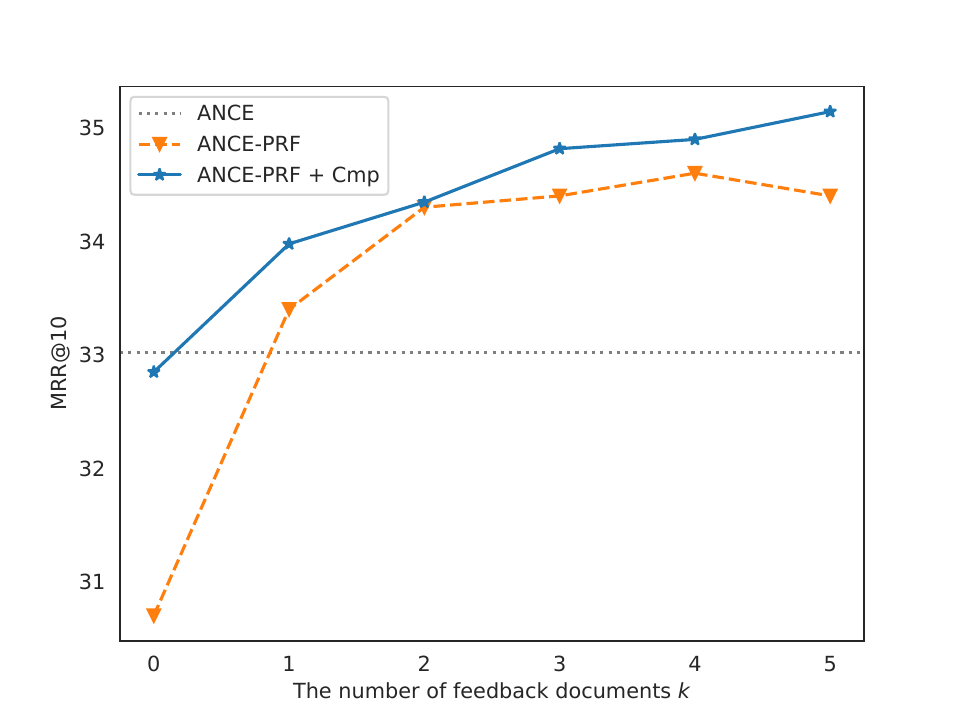}
    \label{fig:prf-ctx}
}
\hfil
\subfigure[]{
    \includegraphics[width=0.5\textwidth]{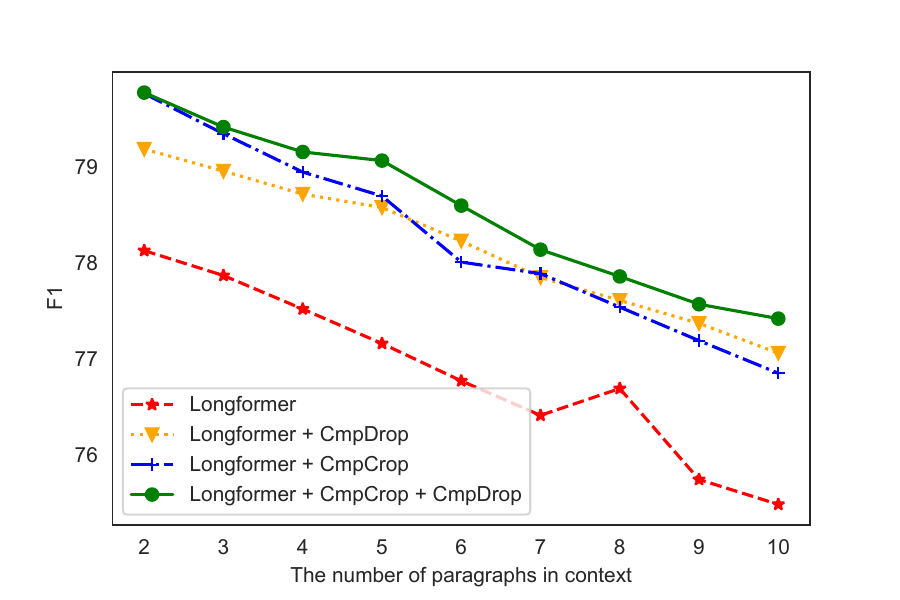}
    \label{fig:rc-ctx}
}
\caption{
Performance curves using different context sizes.
(a) PRF models on MARCO Dev, the horizontal dotted line represents the base retrieval model.
(b) RC models on HotpotQA Dev.
}
\label{fig:ctx-size}
\end{figure}

To review the utility of the input context (i.e., input neurons) to models, we plot in Fig.~\ref{fig:ctx-size} the performance trends of the models using different context sizes.
First, in both datasets, our models trained with comparative loss consistently outperform the baseline models for all context sizes, indicating that our models are able to utilize input neurons more efficiently with equal amounts of input context.
Second, this shows that our comparative loss can further improve the model performance after streamlining the input with context selection.
In addition, we notice that our ANCE-PRF + CmpCrop in Fig.~\ref{fig:prf-ctx} improves retrieval performance as expected as the number of feedback documents increases, while ANCE-PRF reaches peak performance at 4 feedback documents and then suffers performance degradation, implying that our model is more robust and able to mine and exploit relevant information in the added feedback documents.
In contrast to PRF, for HotpotQA in Fig.~\ref{fig:rc-ctx}, the performance of all RC models decreases as the number of paragraphs increases. 
This is understandable, since only 2 paragraphs in HotpotQA are supporting facts, and the remaining 8 mostly serve as a distraction, so the ideal performance curve can just be a horizontal line that does not drop when the paragraph number increases.
Interestingly, we find that the degradation of Longformer + CmpDrop (2.7\%) and Longformer + CmpCrop + CmpDrop (3.0\%) from the oracle setting (2 gold paragraphs) to the distractor setting (10 paragraphs) is lower than that of the baseline Longformer (3.4\%).
This suggests that comparative loss can help the models suppress the noisy information in the added context.
Although Longformer + CmpCrop (3.7\%) has a larger degradation than Longformer, we believe this is because Longformer + CmpCrop needs to be optimized for various numbers of paragraphs, unlike other models without CmpCrop that focus on learning for one input form (i.e., always ten paragraphs). 
However, this variety of input forms makes Longformer + CmpCrop perform better than Longformer + CmpDrop when the number of paragraphs is small ($\le 5$).

To further quantitatively demonstrate the help of comparative loss in the robustness of the PRF model to context size, we report in Table~\ref{tab:prf-ri} the robustness indexes~\cite{collins-thompson2009} of ANCE-PRF + CmpCrop and ANCE-PRF at different numbers of feedback documents.
The robustness index is defined as $\frac{N_{+} - N_{-}}{|Q|}$, where $|Q|$ is the total number of evaluated queries and $N_+$ and $N_-$ are the number of queries that the PRF model improves or downgrades when one more feedback document is used.
The value of robustness index is in [-1, 1], and the model with higher robustness index is more robust.
We can see that the PRF model trained using comparative loss with CmpCrop is significantly more robust than the baseline model.
Besides, from the gaps in their robustness indexes (only 0.03 or 0.02 for 1 or 2 documents, but 0.05 for more documents), we can find that the comparative loss is more helpful for long-form inputs.

\begin{table}
\centering
\caption{The robustness index of $\bm{q}^{(k)}$ with respect to $\bm{q}^{(k-1)}$ on MARCO Dev at each PRF depth $k$, where $\bm{q}^{(k)}$ and $\bm{q}^{(k-1)}$ are reformulated query vectors by the PRF model, the latter having one less document in the input context than the former.}
\label{tab:prf-ri}
\begin{tabular}{lccccc} 
\toprule
\multicolumn{1}{c}{$k$} & 1             & 2             & 3             & 4             & 5              \\ 
\midrule
ANCE-PRF                & 0.51          & 0.54          & 0.58          & 0.58          & 0.61           \\
ANCE-PRF + Cmp (1 Crop) & \textbf{0.54} & \textbf{0.56} & \textbf{0.63} & \textbf{0.63} & \textbf{0.66}  \\
\bottomrule
\end{tabular}
\end{table}

\subsection{Loss Visualization}\label{sec:curves}

\begin{figure}[!t]
\centering
\subfigure[Answer extraction loss on HopotQA Train]{
    \includegraphics[width=0.44\textwidth]{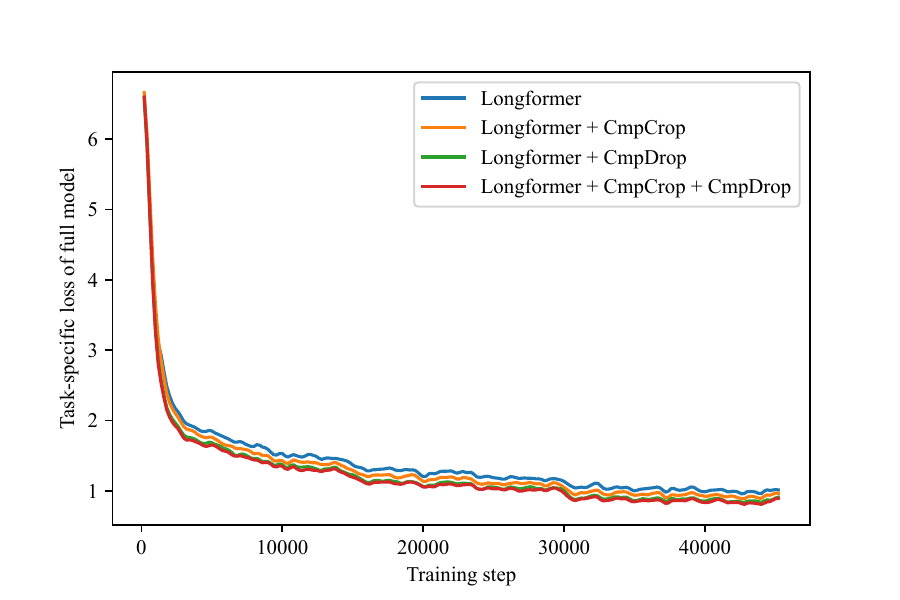}
    \label{fig:qa-train}
}
\hfil
\subfigure[Answer extraction loss on HopotQA Dev]{
    \includegraphics[width=0.44\textwidth]{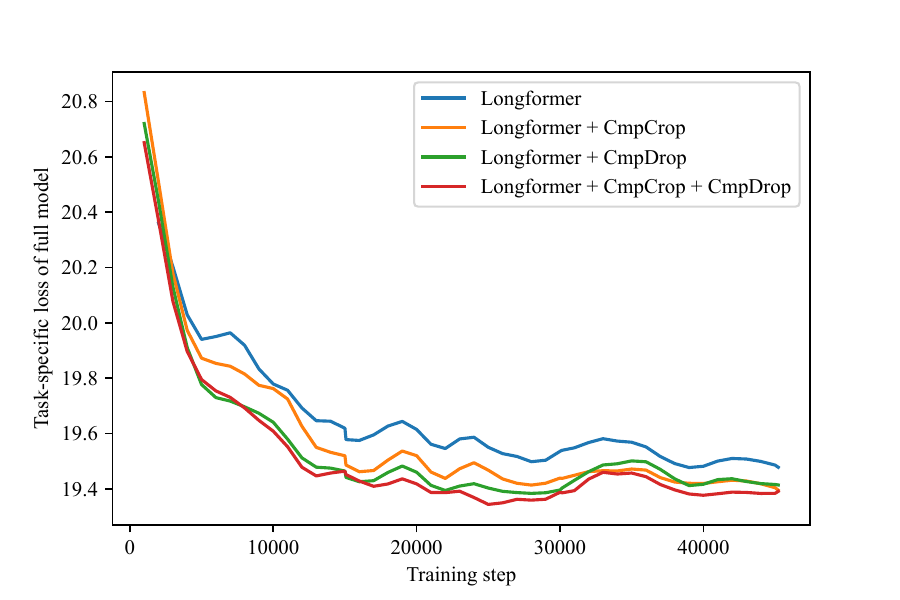}
    \label{fig:qa-dev}
}

\subfigure[Retrieval loss on MARCO Train]{
    \includegraphics[width=0.44\textwidth]{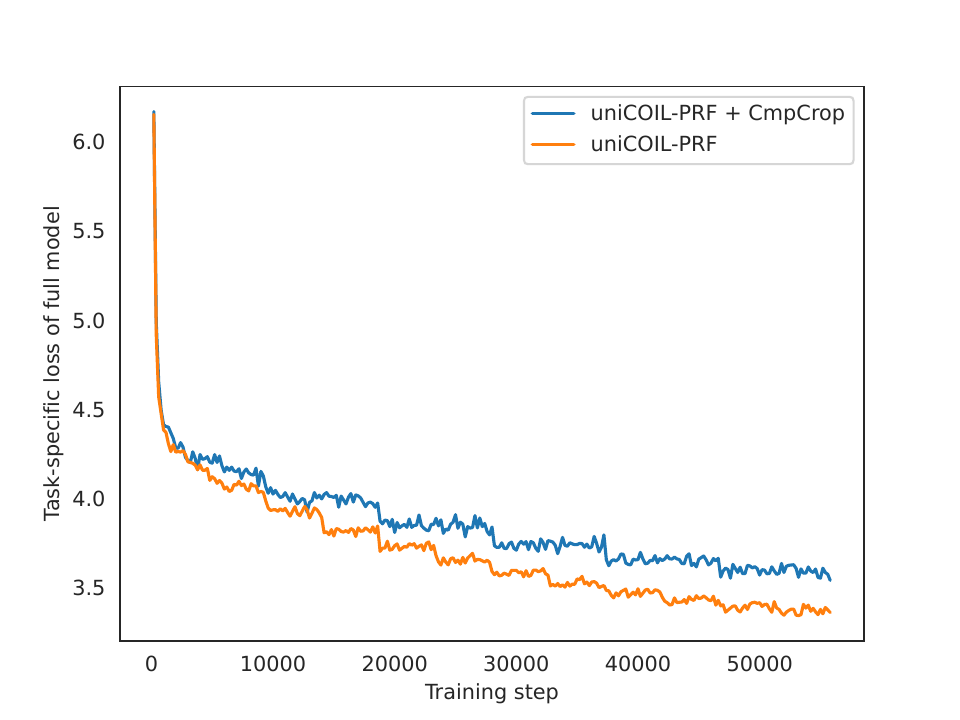}
    \label{fig:prf-train}
}
\hfil
\subfigure[Retrieval loss on MARCO Dev]{
    \includegraphics[width=0.44\textwidth]{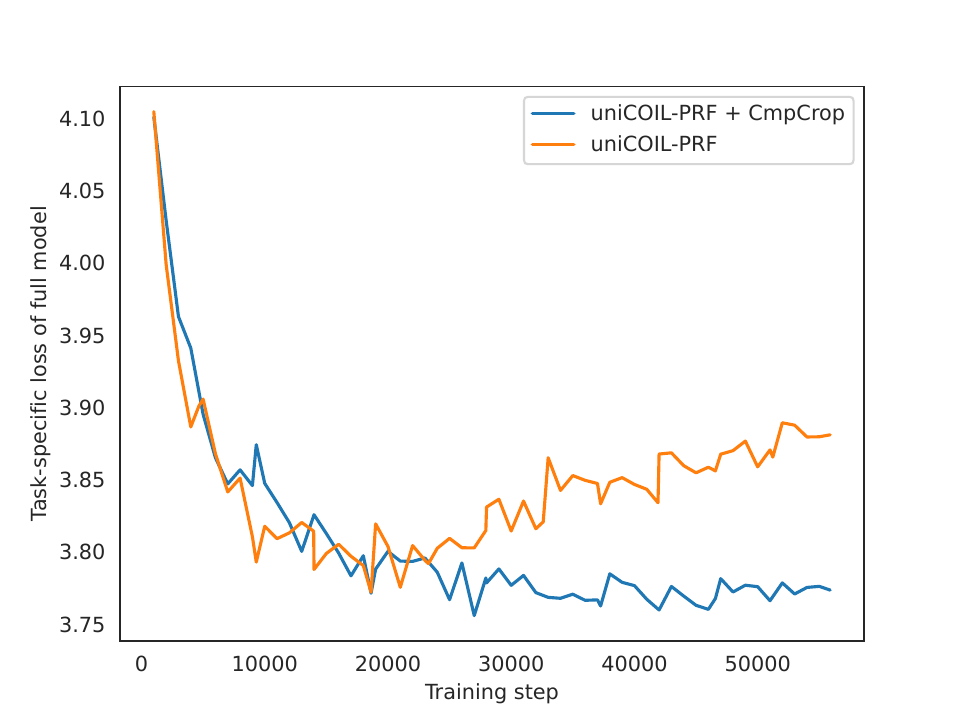}
    \label{fig:prf-dev}
}
\caption{
Task-specific loss curves for the full model.
}
\label{fig:loss-curve}
\end{figure}

To figure out the impact of comparative loss on task-specific loss, we plot the curves of task-specific loss for the full model (i.e., $l^{(0)}$) in Fig.~\ref{fig:loss-curve}.
From Fig.~\ref{fig:qa-train} and Fig.~\ref{fig:qa-dev} we can see that with the same batch size, the comparative loss can help our models fit better compared to the baseline Longformer. 
Comparing Longformer + CmpDrop and Longformer + CmpCrop, we can find that the training loss of the former is significantly smaller, which indicates that the comparative loss with CmpDrop helps the model fit the training data better. 
Whereas the evaluation loss of Longformer + CmpCrop rises less in the later stage, which indicates that the comparative loss with CmpCrop can mitigate the overfitting to some extent.
Since the number of task-specific losses per sample optimized by comparative loss is $1+c$ times that of conventional training, we also plot the task-specific loss curves for PRF models in Fig.~\ref{fig:prf-train} and Fig.~\ref{fig:prf-dev}, where the batch size of our uniCOIL-PRF + CmpCrop is $1/(1+c)$ of the baseline uniCOIL-PRF. 
In this way, the number of task-specific losses optimized in one batch for our model and the baseline is the same, which helps to further clarify the role of the comparative loss with CmpCrop.
We can see that while the training loss of our model in Fig.~\ref{fig:prf-train} does not drop as low as the baseline, its evaluation loss in Fig.~\ref{fig:prf-dev} drops to a lower level and significantly mitigates the overfitting.

\subsection{Training Efficiency}\label{sec:cost}

\begin{table}[!t]
\centering
\caption{Specific settings for the number of ablation steps of BERT + Cmp on each GLUE dataset, as well as the performance gain and increase in training computation overhead compared to BERT.}
\label{tab:efficiency}
\begin{tabular}{lcccccccc} 
\toprule
                            & MNLI & MRPC & QNLI & QQP  & RTE  & SST-2 & STS-B & CoLA  \\ 
\midrule
$c$                         & 3    & 1    & 4    & 2    & 1    & 2     & 4     & 4     \\
Performance (\%) $\uparrow$ & +1.3 & +3.8 & +0.9 & +0.5 & +4.1 & +0.4  & +0.6  & +1.6  \\
FLOPs $\uparrow$            & x3.5 & x1.6 & x3.5 & x0.9 & x2.1 & x0.7  & x4.8  & x3.9  \\
\bottomrule
\end{tabular}
\end{table}

We present in Table~\ref{tab:efficiency} the performance gain and relative change in training FLOPs of BERT$_\mathrm{base}$ + Cmp compared to BERT$_\mathrm{base}$, as well as the specific number of comparisons (i.e., number of ablation steps $c$) chosen for each dataset.
We find that the actual overhead of training with comparative loss is usually less than $1+c$ times that of conventional training, and even less than that of conventional training (e.g., on QQP).
This is because models trained with comparative loss tend to converge earlier than baselines.
Combined with the insensitivity of comparative loss to the number of comparisons found from Fig.~\ref{fig:num-drop}, we believe that setting $c$ to 1 or 2 can lead to effective and fast training when data is sufficient.

\section{Related Work}
In this section, we introduce and discuss some work that has different motivations but is technically relevant to us, starting with contrastive learning~\cite{le-khac2020} that learns by comparing, followed by recent training methods that also use dropout multiple times.

\subsection{Contrastive Learning}
Contrastive learning has recently achieved significant success in representation learning in computer vision and natural language processing.
At its core, contrastive learning aims to learn effective representations by pulling semantically similar neighbors together and pushing apart non-neighbors~\cite{hadsell2006dimensionality}.
Instead of learning a signal from individual data samples one at a time, it learns by comparing different samples~\cite{le-khac2020}. 
The comparison is performed between positive pairs of similar samples and negative pairs of dissimilar samples.
The positive pair must ensure that the two samples are similar, which can be constructed either by using supervised similarity annotation or by self-supervision.
In self-supervised contrastive learning, a positive pair can consist of an original sample and its data augmentation.
For example, SimCLR~\cite{chen2020simclr} in computer vision uses a crop, flip, distortion or rotation of an original image as its similar view, and SimCSE~\cite{gao2021simcse} in natural language processing applies two dropout masks to an input sentence to create two slightly different sentence embeddings that are then used as a positive pair of sentence embeddings.
To share more computation and save cost, negative pairs usually consist of two dissimilar samples within the same training batch.
Although both learn through comparison, contrastive learning aims at pursuing alignment and uniformity~\cite{wang2020a} of representations, while our comparative loss aims at pursuing orderliness of the task-specific losses of the full model and its ablated model.
Moreover, as the lexical meaning suggests, contrastive learning only classifies the relationship (i.e., similar or dissimilar) between two data samples in a binary manner, whereas our comparative loss compares multiple full/ablated models by ranking.
However, these two are not in conflict, and our comparative loss can be used over the contrastive losses that served as task-specific losses.

\subsection{Dropout-based Comparison}
Dropout is a family of stochastic techniques used in neural network training or inference that have attracted extensive research interest and are widely used in practice. 
The standard dropout~\cite{hinton2012dropout} aims to avoid overfitting of the network by reducing the co-adaptation of neurons, where the outputs of individual neurons only provide useful information in combination with other neuron outputs. 
After this, a line of research focused on improving the standard dropout by employing other strategies for dropping neurons, such as dropconnect~\cite{wan2013dropconnect} and variational dropout~\cite{kingma2015}.

A line of research that is relevant to us is the use of dropout multiple times in training.
SimCSE~\cite{gao2021simcse} forwards the model twice with different dropout masks of the same rate and uses a contrastive loss to constrain the distribution of model outputs in the representation space.
A possible side effect of dropout revealed by the existing literature~\cite{ma2016,zolna2017} is the non-negligible inconsistency between the training and inference stages of the model, i.e., the submodels are optimized during training, but the full model without dropout is used during inference.
To address this inconsistency, R-Drop~\cite{liang2021} forward runs the model multiple times with different dropout masks to obtain multiple predicted probability distributions and applies KL-divergence on them to constrain their consistency.
Unlike their multiple dropout masks that are sampled independently, the multiple dropout rates are increasing and the masks are progressive in our CmpDrop, with the subsequent mask obtained by further randomly discarding elements based on the previous one.
In addition, we impose constraints on the task-specific losses at the end rather than on the representations and probabilities upstream.
Notably, the full model is also optimized in due time when trained using the comparative loss with CmpDrop, which we argue is important to mitigate the inconsistency between training and inference.
This is because, while dropout avoids co-adaptation of neurons, it also weakens the cooperation between neurons ($\S$\ref{sec:weighting} gives some empirical support).
In particular, in cases where all neurons are involved, the full model trained with dropout has not been taught how to make them work together efficiently and thus cannot be fully exploited during testing.
Surprisingly, our comparative loss with CmpDrop can balance between promoting the cooperation of neurons and preventing their co-adaptation.

\section{Conclusion}
In this paper, we propose cross-model comparative loss, a simple task-agnostic loss function, to improve the utility of neurons in NLU models.
Comparative loss is essentially a ranking loss based on the comparison principle between the full model and its ablated models, with the expectation that the less ablation there is, the smaller the task-specific loss.
To ensure comparability among multiple ablated models, we progressively ablate the models and provide two controlled ablation methods based on dropout and context cropping, applicable to a wide range of tasks and models.
We show theoretically how comparative loss works, suggesting that it can adaptively assign weights to multiple task-specific losses.
Extensive experiments and analysis on 14 datasets from 3 distinct NLU tasks demonstrate the universal effectiveness of comparative loss.
Interestingly, our analysis confirms that comparative loss can indeed assign weights more appropriately, and finds that comparative loss is particularly effective for models with few parameters or long input.

In the future, we would like to apply comparative loss in other domains, such as natural language generation and computer vision, and explore its applications on other model architectures beyond Transformer.
It could also be interesting to explore the application of comparative loss on top of self-supervised losses (e.g., contrastive loss) during pretraining.
For training costs, how to reduce the overhead by reusing more shared computations is a direction worth exploring.
Further, more advanced ablation methods in training, such as dropconnect~\cite{wan2013dropconnect} rather than standard dropout and adversarial rather than stochastic, may deserve future research efforts.

\begin{acks}
This work was supported by the  \grantsponsor{GS1}{National Key R\&D Program of China}{} (\grantnum{GS1}{2022YFB3103700}, \grantnum{GS1}{2022YFB3103704}), the \grantsponsor{GS2}{National Natural Science Foundation of China (NSFC)}{} under Grants No. \grantnum{GS2}{62276248}, \grantnum{GS2}{U21B2046}, and the \grantsponsor{GS3}{Youth Innovation Promotion Association CAS}{} under Grants No. \grantnum{GS3}{2023111}.
\end{acks}


\bibliographystyle{ACM-Reference-Format}
\bibliography{main}


\end{document}